\documentclass[pdflatex,sn-mathphys-num]{sn-jnl}

\usepackage{graphicx}%
\usepackage{multirow}%
\usepackage{float}%
\usepackage{amsmath,amssymb,amsfonts}%
\usepackage{amsthm}%
\usepackage{mathrsfs}%
\usepackage[title]{appendix}%
\usepackage[table]{xcolor}%
\usepackage{textcomp}%
\usepackage{manyfoot}%
\usepackage{booktabs}%
\usepackage{algorithm}%
\usepackage{algorithmicx}%
\usepackage{algpseudocode}%
\usepackage{listings}%
\usepackage{bm}%
\usepackage{pifont}%
\usepackage{dsfont}%
\usepackage{url}%
\usepackage[caption=false,font=normalsize,labelfont=sf,textfont=sf]{subfig}

\definecolor{Gray}{gray}{0.9}
\definecolor{mygreen}{HTML}{39b54a}
\newcommand{\reshl}[2]{
{#1} \fontsize{7.5pt}{1em}\selectfont\color{mygreen}{$\uparrow$ \textbf{#2}}
}

\theoremstyle{thmstyleone}%

\theoremstyle{thmstyletwo}%

\theoremstyle{thmstylethree}%

\let\table\tableorg
\let\endtable\endtableorg

\begin{document}

\title[Hyperspectral Smoke Segmentation]{Hyperspectral Smoke Segmentation \hb via Mixture of Prototypes}

\author{\fnm{Lujian} \sur{Yao}}\email{lujianyao@mail.ecust.edu.cn}

\author*{\fnm{Haitao} \sur{Zhao\textsuperscript{*}}}\email{haitaozhao@ecust.edu.cn}

\author{\fnm{Xianghai} \sur{Kong}}\email{xianghaikong35@mail.ecust.edu.cn}

\author{\fnm{Yuhan} \sur{Xu}}\email{yuhanxu@mail.ecust.edu.cn}

\affil{\orgdiv{Automation Department, School of Information Science and Engineering}, \orgname{East China University of Science and Technology}, \orgaddress{\street{Meilong Rd. No.130}, \city{Shanghai}, \postcode{200237}, \country{China}}}

\abstract{Smoke segmentation is critical for wildfire management and industrial safety applications. Traditional visible-light-based methods face limitations due to insufficient spectral information, particularly struggling with cloud interference and semi-transparent smoke regions. To address these challenges, we introduce hyperspectral imaging for smoke segmentation and present the first hyperspectral smoke segmentation dataset (HSSDataset) with carefully annotated samples collected from over 18,000 frames across 20 real-world scenarios using a \textit{Many-to-One annotations} protocol. 
However, different spectral bands exhibit varying discriminative capabilities across spatial regions, necessitating adaptive band weighting strategies. We decompose this into three technical challenges: spectral interaction contamination, limited spectral pattern modeling, and complex weighting router problems.
We propose a mixture of prototypes (MoP) network with: (1) band split (BS) for spectral isolation, (2) prototype-based spectral representation (PSR) for diverse patterns, and (3) dual-stage router (DSR) for adaptive spatial-aware band weighting.
We further construct a multispectral dataset (MSSDataset) with RGB-infrared images. Extensive experiments validate superior performance across both hyperspectral and multispectral modalities, establishing a new paradigm for spectral-based smoke segmentation.}


\maketitle

\section{Introduction}\label{sec1}

Smoke segmentation plays a crucial role in both wildfire management and industrial safety applications through its ability to precisely localize smoke presence. In wildfire scenarios, smoke segmentation serves as an early warning indicator, enabling rapid identification of fire sources and facilitating timely emergency response~\cite{zervas2011multisensor}. In industrial settings, accurate smoke segmentation helps detect gas leaks and hazardous emissions early, preventing potential accidents and enabling swift containment measures~\cite{utomo2026explainable}. The broad applicability and critical safety implications make smoke segmentation an important research area with significant real-world impact.

Traditional smoke segmentation approaches primarily focused on analyzing specific smoke characteristics, including color space~\cite{xing2015smoke}, textural analysis, and color enhancement strategies~\cite{jia2016saliency}. Vision-based techniques such as morphological operations~\cite{filonenko2017fast}, and transmission estimation~\cite{long2010transmission} are extensively employed.  However, these traditional methods struggled with the variability of smoke and often required manual parameter tuning for different environmental conditions and scenarios.
Deep learning-based smoke segmentation approaches have made significant progress through several key strategies. Multi-scale feature fusion~\cite{yuan2021gated,yuan2022cubic} integrates features from multiple levels to capture both local details and global context, while receptive field expansion~\cite{jing2023smokeseger,li20183d} enables better understanding of smoke's spatial extent and morphological variations. Progressive segmentation~\cite{yuan2019deep,zhang2021att} employs coarse-to-fine refinement to iteratively improve boundary delineation, and uncertainty quantification~\cite{yan2022transmission} helps handle ambiguous regions where smoke boundaries are unclear.

However, visible-light-based methods face critical limitations due to the narrow spectral range of visible light bands, which provides insufficient spectral information to distinguish smoke from visually similar phenomena. Similar to camouflaged object detection~\cite{song2025continuous}, where objects blend seamlessly with their surroundings, smoke segmentation also faces the challenge of detecting semi-transparent and low-contrast regions that blend with the background. Performance is easily constrained by {two key challenges}: \textbf{(1) Cloud interference:} Clouds and smoke exhibit similar visual characteristics, making it difficult to distinguish them in visible light images. \textbf{(2) Semi-transparent smoke regions:} Semi-transparency commonly appears along smoke edges with spatially varying opacity, low contrast against the background. These properties produce diffuse boundaries, leading to unclear transitions and inconsistent visual patterns that confuse segmentation models.

    \begin{figure}[t]
        \begin{center}
        \includegraphics[width=\linewidth]{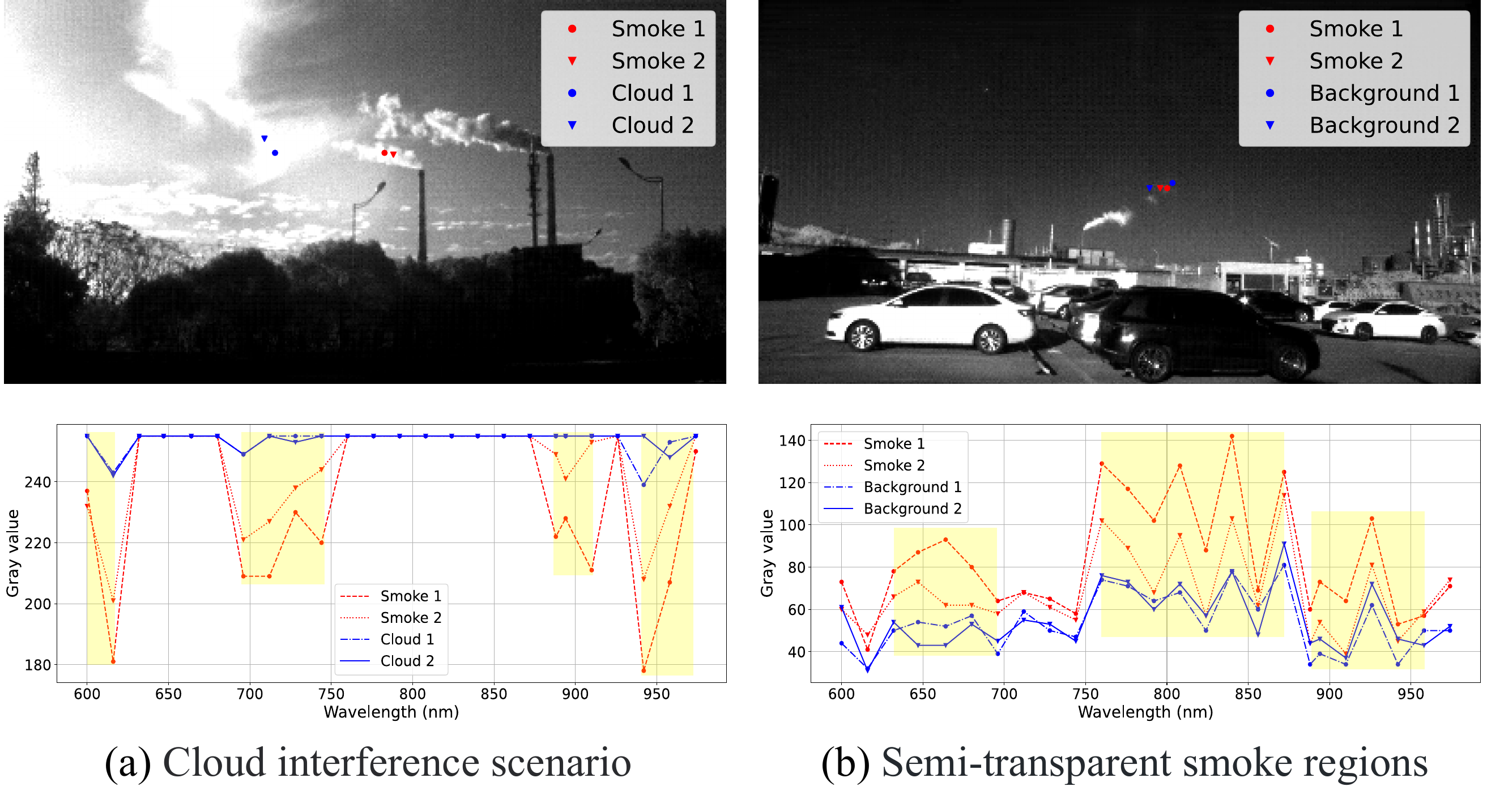}
    \end{center}
    \caption{Motivation for hyperspectral smoke segmentation. The upper part shows challenging smoke scenarios with cloud interference and semi-transparent regions in the visible light band. The lower part plots the spectral distribution of the marked points. Yellow shaded regions highlight the key discriminatory band ranges where smoke and clouds (or smoke and background) exhibit the most significant spectral divergence.}
        \label{fig:motivation}
    \end{figure}

To address the limitations of visible light data, we introduce hyperspectral imaging for smoke segmentation. Hyperspectral imaging captures spectral information across several narrow, contiguous spectral bands, typically spanning the visible, near-infrared, and shortwave infrared regions of the electromagnetic spectrum. 
Each pixel in a hyperspectral image contains a spectral signature, forming a unique ``spectral fingerprint'' that characterizes the material composition and physical properties of the corresponding ground object. This spectral information is particularly valuable for smoke segmentation, as smoke particles exhibit distinct absorption and scattering properties across different wavelengths due to their size distribution, chemical composition, and temperature characteristics. The fine spectral resolution enables detection of subtle spectral variations that are invisible to conventional visible-light cameras, making it possible to differentiate between smoke and spectrally similar phenomena.

For example, in Figure~\ref{fig:motivation}(a), smoke and clouds appear nearly indistinguishable in visible light due to their similar white appearance. However, examining different hyperspectral bands reveals clear distinctions between smoke and clouds in certain spectral ranges. In Figure~\ref{fig:motivation}(b), semi-transparent smoke regions that are difficult to differentiate from the background in visible light exhibit distinct spectral signatures across hyperspectral bands, enabling more reliable segmentation. These examples highlight how hyperspectral imaging provides critical discriminative information that is unavailable in conventional visible light images.

To promote this research area, we introduce the first hyperspectral smoke segmentation dataset (HSSDataset), built from an extensive collection of over 18,000 hyperspectral video frames captured across 20 real-world smoke scenarios using a hyperspectral camera with 25 spectral bands (600-974nm). From this large-scale raw data collection, we carefully selected and annotated 1,007 high-quality samples under diverse challenging conditions, including high exposure, low visibility, early-stage minimal smoke, cloud interference, and complex backgrounds. To ensure annotation reliability and capture the inherent uncertainty in smoke boundary delineation, we introduce a rigorous \textit{Many-to-One annotations} protocol, where each frame receives three independent annotations from different annotators, and the final ground truth is determined through majority voting consensus.

However, directly applying hyperspectral data to smoke segmentation faces a fundamental challenge: \textbf{Different spectral bands have varying discriminative capabilities in different spatial regions.} 
For instance, some bands excel at capturing smoke transparency while others are more effective at distinguishing smoke from visually similar objects like clouds or fog. 
This regional-spectral variation necessitates adaptive \textit{band weighting} strategies that can dynamically assign different spectral band weighting to different spatial regions. This variation creates three main technical challenges:
\textbf{\ding{182} Spectral interaction contamination:} Simply employing convolution or transformer-based backbones can mix spectral channels, causing spectral interactions that contaminate the spectral signatures. Mixing bands with different discriminative purposes dilutes their respective capabilities, making it difficult to leverage band-specific characteristics. This issue is well-recognized in band selection literature~\cite{sun2019hyperspectral,li2025hyperspectral}, where not all spectral bands contribute equally and redundant or noisy bands can degrade performance. Our ablation study (Table~\ref{tab:ablation} and Figure~\ref{fig:band_split}) further validates the effectiveness of spectral isolation.
\textbf{\ding{183} Limited spectral pattern modeling capability:} Conventional approaches rely on monolithic feature encoders that cannot adequately capture the complex and diverse spectral patterns present in hyperspectral data. The diversity of spectral patterns requires specialized representation learning for spectral patterns characterized by multiple distribution centers, where different spectral bands may exhibit distinct clustering patterns in the feature space, reflecting the diverse material compositions and physical properties of smoke across different wavelength ranges.
\textbf{\ding{184} Weighting router problem:} Given the presence of spectral interaction contamination and the diversity of spectral patterns, determining optimal band weighting becomes highly complex. The weighting mechanism must adapt to varying input conditions and dynamically adjust weighting across different spatial regions to capture the intricate dependencies between spectral characteristics and spatial context. 

To address these challenges, we propose a mixture of prototypes (MoP) network that leverages \textit{prototype learning} to adaptively model diverse spectral patterns across hyperspectral bands. The core principle of MoP is to enable dynamic band weighting that automatically adapts to input-specific feature characteristics, allowing the model to emphasize the most discriminative spectral bands. 
{Prototype learning} is a representation learning approach where multiple prototype vectors serve as representative templates that capture different data patterns or categories.

Specifically, the MoP addresses these challenges through three key innovations: \textbf{(For \ding{182}) band split (BS)} processes each spectral band independently through dedicated neural network branches, ensuring complete spectral isolation to preserve spectral signatures and eliminate interaction contamination. \textbf{(For \ding{183}) prototype-based spectral representation (PSR)} employs multiple specialized prototypes to capture diverse spectral patterns, with each prototype learning to encode specific spectral characteristics rather than using uniform representations.
\textbf{(For \ding{184}) dual-stage router (DSR)} addresses the weighting router problem with a two-stage mechanism that dynamically adjusts band weighting across spatial regions under varying input conditions. It first aggregates prototypes via the prototype router (P-Router) to capture intricate dependencies between spectral characteristics, then employs the feature router (F-Router) to produce spatial-aware band weighting.

To demonstrate the generalizability of our MoP, we further extend our validation beyond hyperspectral data by constructing a multispectral smoke segmentation dataset (MSSDataset) that contains RGB--infrared
paired images. This MSSDataset comprises carefully annotated smoke samples across diverse scenarios.
Our MoP achieves consistent superior performance on both hyperspectral and multispectral data.

Our main contributions can be summarized as follows:

\begin{itemize}
    \item We introduce the first hyperspectral smoke segmentation task and establish a comprehensive hyperspectral smoke segmentation dataset (HSSDataset), employing a rigorous \textit{Many-to-One annotations} protocol to ensure reliable ground truth.
    \item A novel mixture of prototypes (MoP) network is proposed that performs spatial-aware band weighting on input-specific features across diverse smoke scenarios.
    \item We develop three key innovations within the MoP: BS for spectral isolation to preserve spectral signatures, PSR for diverse pattern modeling, and DSR for adaptive weight allocation.
    \item To validate the generalizability of our MoP, we construct a complementary multispectral smoke segmentation dataset (MSSDataset) and demonstrate consistent superior performance across both hyperspectral and multispectral modalities.
\end{itemize}

\section{Related Work}\label{sec2}
\subsection{Smoke Segmentation}

Traditional smoke segmentation approaches primarily focused on analyzing specific smoke characteristics, including color space transformations~\cite{xing2015smoke}, textural analysis, and color enhancement strategies~\cite{jia2016saliency}. Vision-based techniques such as morphological operations~\cite{filonenko2017fast}, and transmission estimation~\cite{long2010transmission} are extensively employed. While computationally efficient and easy to implement, these traditional methods struggled with the inherent variability of smoke and often required manual parameter tuning for different environmental conditions and scenarios.

Deep learning-based smoke segmentation approaches address the aforementioned challenges through several key strategies: 
{(1) Multi-scale feature fusion:} Methods such as~\cite{yuan2021gated,yuan2022cubic} integrate features from multiple abstraction levels to capture both local details and global context, essential for handling the multi-scale nature of smoke. 
{(2) Receptive field expansion:} Techniques including~\cite{jing2023smokeseger,li20183d} augment the perceptual field of neural networks to better capture the spatial extent of smoke and its morphological variations. By expanding the receptive field, these methods can better understand the global context and long-range dependencies that are crucial for accurate smoke detection.
{(3) Coarse-to-fine refinement:} Progressive segmentation strategies~\cite{yuan2019deep,zhang2021att} employ hierarchical processing to refine segmentation boundaries iteratively. These methods typically start with coarse predictions and progressively refine the results to achieve more precise boundary delineation.
{(4) Uncertainty quantification:} Advanced methods~\cite{yan2022transmission} incorporate uncertainty estimation mechanisms to handle ambiguous regions and improve robustness in challenging scenarios where smoke boundaries are inherently unclear.

However, visible-light-based smoke segmentation methods face critical limitations due to the narrow spectral range of visible light bands, which provides insufficient discriminative information for complex scenarios.

\subsection{Hyperspectral Image}

Hyperspectral imaging captures data across hundreds of narrow, contiguous spectral bands~\cite{Ahmad2022hyperspectral,hong2020graph}, forming a 3D data cube $(h, w, d)$ where each pixel contains a rich spectral signature that characterizes material composition and physical properties~\cite{Du2016beyond,sun2019spectral,Pande2022hyperloopnet}.
This rich spectral information makes hyperspectral imaging particularly suitable for challenging smoke segmentation tasks, where conventional visible-light-based data struggles with cloud interference, varying transparency, and complex backgrounds.

Recent advances in hyperspectral image processing have explored diverse strategies for spectral-spatial feature extraction and fusion~\cite{feng2025fractional}. Our MoP differs from these generic approaches in two key aspects: (1) Existing methods such as spatial-spectral cross-fusion networks~\cite{fu2025gfhmp} and asymptotic spectral mapping~\cite{liu2024asymptotic} typically treat all spectral bands uniformly or apply fixed band selection, whereas our MoP enables \textit{dynamic} and \textit{spatial-aware} band weighting through band-level prototypes that adapts to input-specific features. (2) Methods like multi-source spectral data combination~\cite{yao2025combination} and heterospectral structure compensation~\cite{liu2025heterospectral} rely on monolithic encoders for spectral processing, while our band split (BS) strategy ensures complete spectral isolation before adaptive weighting.

\subsection{Prototype Learning and Mixture of Experts}
Prototype learning originated from classical machine learning methods like nearest neighbor algorithms~\cite{cover1967nearest} and cognitive science theories on prototype representation~\cite{knowlton1993learning}. At its core, prototype learning uses representative examples to model different classes, enabling classification by comparing samples to these prototypes. This approach led to various non-parametric classification methods such as learning vector quantization (LVQ)~\cite{kohonen1990self} and neighborhood component analysis (NCA)~\cite{goldberger2004neighbourhood}. Modern deep learning has embraced prototype learning by incorporating it into supervised~\cite{mettes2019hyperspherical}, unsupervised~\cite{wu2018unsupervised}, and self-supervised~\cite{caron2020unsupervised} frameworks, providing structured embedding spaces through multi-centric modeling.

Mixture of Experts (MoE)~\cite{eigen2013learning} involves training multiple specialized networks (experts) alongside a gating mechanism that determines which experts to activate for each input, allowing selective computation and specialized processing. Core MoE approaches include GShard~\cite{lepikhin2020gshard} for transformer-based architectures and Switch Transformer~\cite{fedus2022switch} for sparse expert routing. 

In our work, we integrate prototype learning with MoE to present a natural and promising direction for hyperspectral smoke segmentation. Prototype learning provides interpretable spectral representations as reference benchmarks, while MoE enables efficient processing of diverse spectral patterns through specialized router networks. This combination allows our approach to maintain multiple specialized prototypes for different smoke characteristics while routing spectral features to the most appropriate expert.

\section{Hyperspectral Smoke Segmentation Dataset (HSSDataset)}\label{sec3}

This section provides an overview of our dataset construction process, beginning with the hyperspectral imaging background (Sec.~\ref{sec:hsi_background}), followed by the hyperspectral camera system specifications and data collection methodology (Sec.~\ref{sec:dataset_camera}). We then detail our annotation protocol and quality control measures (Sec.~\ref{sec:dataset_annotation}). Finally, we present the multispectral dataset extension for MoP validation (Sec.~\ref{sec:mssdataset}). Both the HSSDataset and MSSDataset are publicly available on Hugging Face\footnote{Dataset: \url{https://huggingface.co/datasets/LujianYao/HSSDataset}}.

\subsection{Hyperspectral Imaging Background}\label{sec:hsi_background}

Hyperspectral imaging technology has been widely applied in remote sensing and computer vision. Unlike RGB imaging, which compresses the continuous spectrum into three broad bands (red, green, blue), hyperspectral sensors acquire data across tens to hundreds of narrow, contiguous spectral bands covering the visible, near-infrared, and shortwave infrared regions~\cite{Ahmad2022hyperspectral}. A hyperspectral image can be represented as a 3D data cube of dimensions $(h, w, d)$, where $h$ and $w$ are the spatial dimensions and $d$ is the number of spectral bands~\cite{Du2016beyond}. Each pixel contains a continuous spectral response curve, forming a unique ``spectral fingerprint.'' The spectral resolution of hyperspectral data is typically on the order of nanometers ($\sim$5--10\,nm). This fine spectral resolution provides two key advantages: (1) the ability to precisely distinguish materials with similar appearances but different spectral characteristics, and (2) relatively stable spectral features under different illumination conditions, reducing sensitivity to environmental changes.

\begin{figure*}[t]
    \begin{center}
        \includegraphics[width=\linewidth]{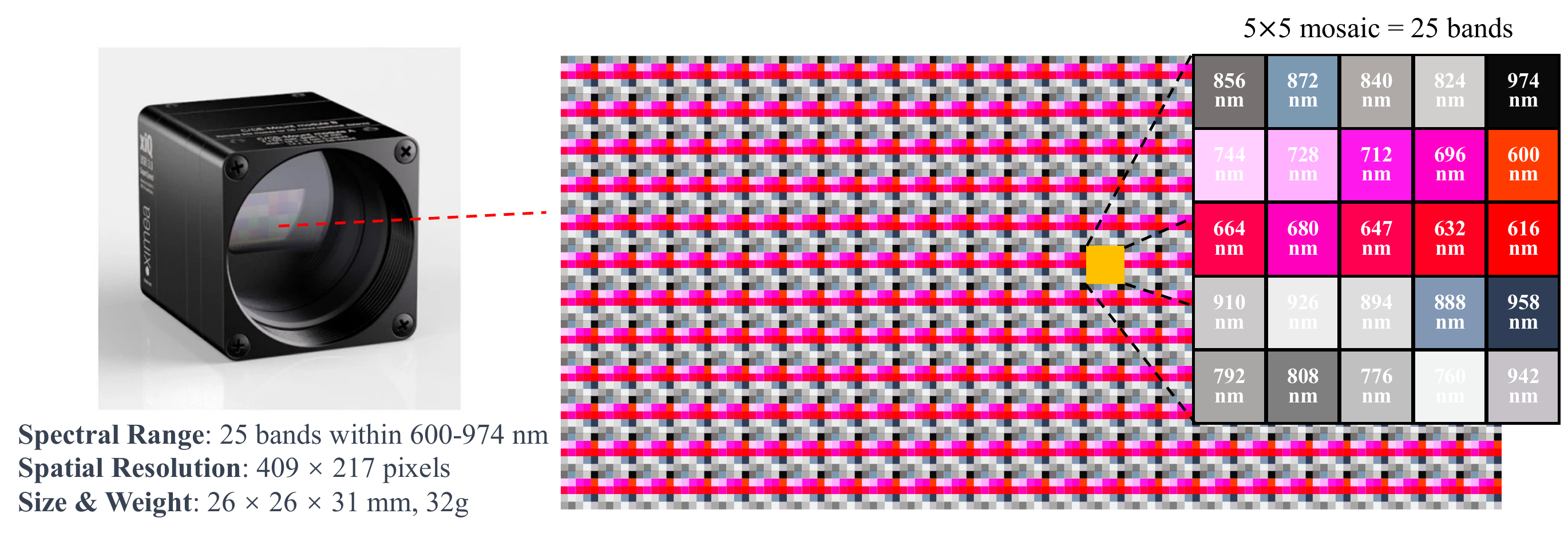}
    \end{center}
    \caption{XIMEA MQ022HG-IM-SM5X5-NIR hyperspectral camera specifications and 25-band mosaic filter design. The camera employs a specialized $5 \times 5$ filter array with wavelengths spanning 600-974nm for simultaneous spatial and spectral information.}
    \label{fig:camera}
\end{figure*}

\subsection{Hyperspectral Camera System and Data Collection}\label{sec:dataset_camera}

We employ a XIMEA MQ022HG-IM-SM5X5-NIR snapshot hyperspectral camera, as shown in Figure~\ref{fig:camera}. This camera combines a high-sensitivity CCD and sCMOS chip with a specialized 25-band mosaic filter coating deployed as a $5 \times 5$ filter array on the sensor surface. Each $5 \times 5$ pixel block simultaneously acquires spectral information across 25 distinct bands, with wavelengths spanning 600--974\,nm: 600, 616, 632, 647, 664, 680, 696, 712, 728, 744, 760, 776, 792, 808, 824, 840, 856, 872, 888, 894, 910, 926, 942, 958, and 974\,nm.

Compared to traditional push-broom hyperspectral cameras, this snapshot imaging approach effectively balances spectral and spatial resolution, requiring only a single exposure to acquire all bands simultaneously. This enables rapid acquisition of both spectral and spatial information, meeting the requirements of dynamic scene capture. Additionally, the highly integrated architecture significantly reduces the camera's size and weight, facilitating flexible deployment.

\begin{figure}[t]
    \centering
    \includegraphics[width=\linewidth]{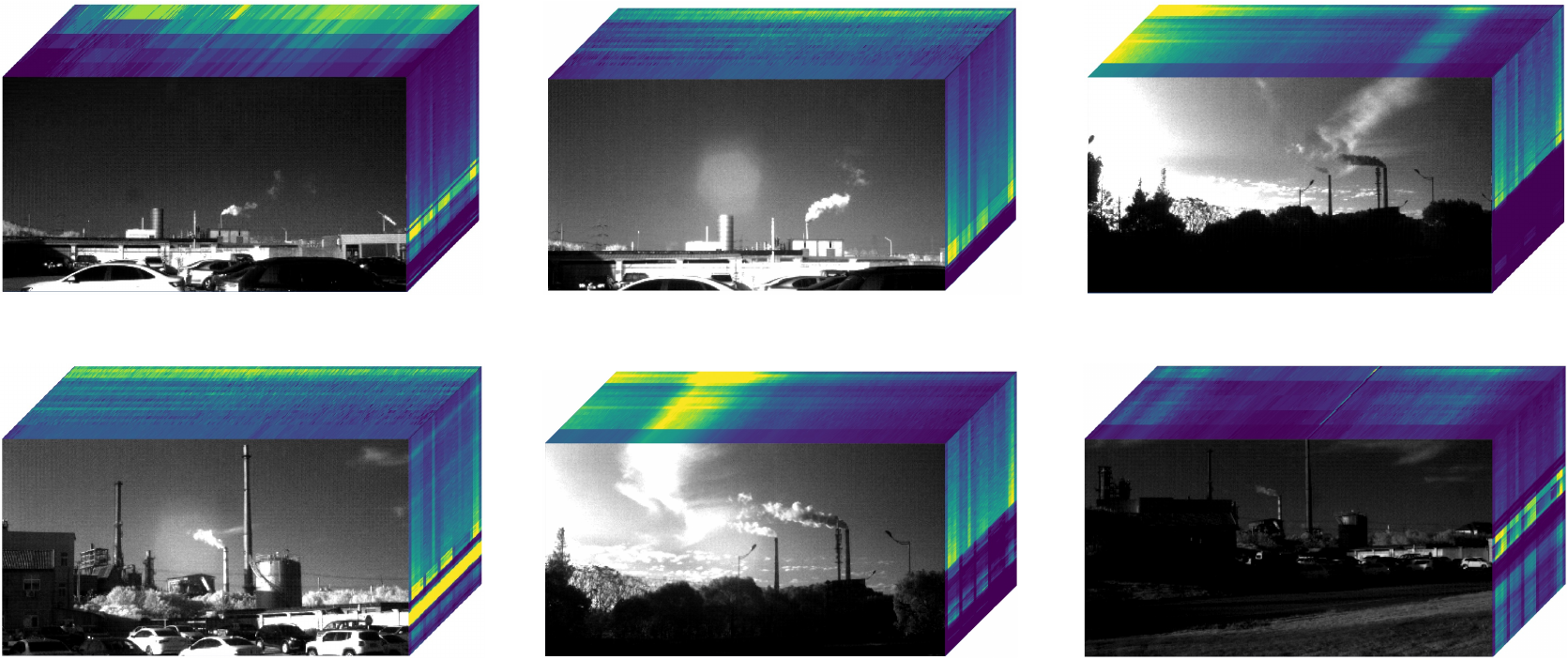}
    \caption{The challenging scenarios of HSSDataset.}
    \label{fig:hss_dataset}
\end{figure}

As shown in Figure~\ref{fig:hss_dataset}, we conduct data collection across 20 real-world industrial emission scenarios, capturing over 18,000 hyperspectral video frames. The collection targeted diverse smoke conditions, including challenging scenarios such as high exposure environments, low visibility conditions, and early-stage minimal smoke detection. We also addressed environmental interference from cloud and complex background elements, and varying atmospheric conditions. The dataset encompasses various smoke characteristics, including semi-transparent regions, blurred boundaries, and varying opacity and particle density across different industrial contexts. Table~\ref{tab:scenario_distribution} provides a comprehensive breakdown of our 1,007 annotated samples across different challenging scenarios. The distribution covers both scene-based classifications and smoke-based classifications.

\begin{table}[t]
\centering
\caption{Distribution of annotated samples across different challenging scenarios in HSSDataset.}
\label{tab:scenario_distribution}
\resizebox{\linewidth}{!}{
\begin{tabular}{l|c|l}
\toprule[1pt]
\textbf{Scenario Category} & \textbf{Sample Count} & \textbf{Description} \\
\midrule
\multicolumn{3}{c}{\textbf{Scene-based Classification}} \\
\midrule
High Exposure & 214 & Bright lighting conditions with overexposed regions \\
Low Visibility & 118 & Poor lighting and atmospheric conditions \\
Complex Background & 411 & Industrial environments with cluttered backgrounds \\
Cloud Interference & 264 & Scenes with cloud-smoke confusion scenarios \\
\midrule
\multicolumn{3}{c}{\textbf{Smoke-based Classification}} \\
\midrule
Early-stage Smoke & 268 & Small smoke plumes in initial emission phases \\
High Transparency Smoke & 258 & Varying smoke opacity and transparency \\
Complex-shaped Smoke & 481 & Irregular smoke patterns with unclear boundaries \\
\bottomrule[1pt]
\end{tabular}}
\end{table}

\subsection{Dataset Annotation Protocol}\label{sec:dataset_annotation}

\noindent\textbf{Sampling Strategy.} From the 18,000+ collected frames, we systematically sample every 18th frame to capture smoke evolution dynamics while maintaining annotation efficiency. This strategy ensures comprehensive coverage of smoke generation, diffusion, and dissipation phases.

\noindent\textbf{Band-Averaged Image Generation.} Since hyperspectral data contains multiple bands that cannot be directly visualized as a single image for annotation, we generate band-averaged grayscale images by computing the arithmetic mean across all spectral bands:
\begin{equation}
\bm{I}_{\text{avg}} = \frac{1}{d}\sum_{\ell=1}^{d}\bm{I}^{(\ell)},
\end{equation}
where $\bm{I}^{(\ell)}$ denotes the $\ell$-th band image and $d=25$ is the total number of bands. This approach creates intuitive visual representations that facilitate human interpretation while preserving the overall spatial structure and intensity patterns of the smoke regions.

\begin{figure}[t]
    \begin{center}
        \includegraphics[width=\linewidth]{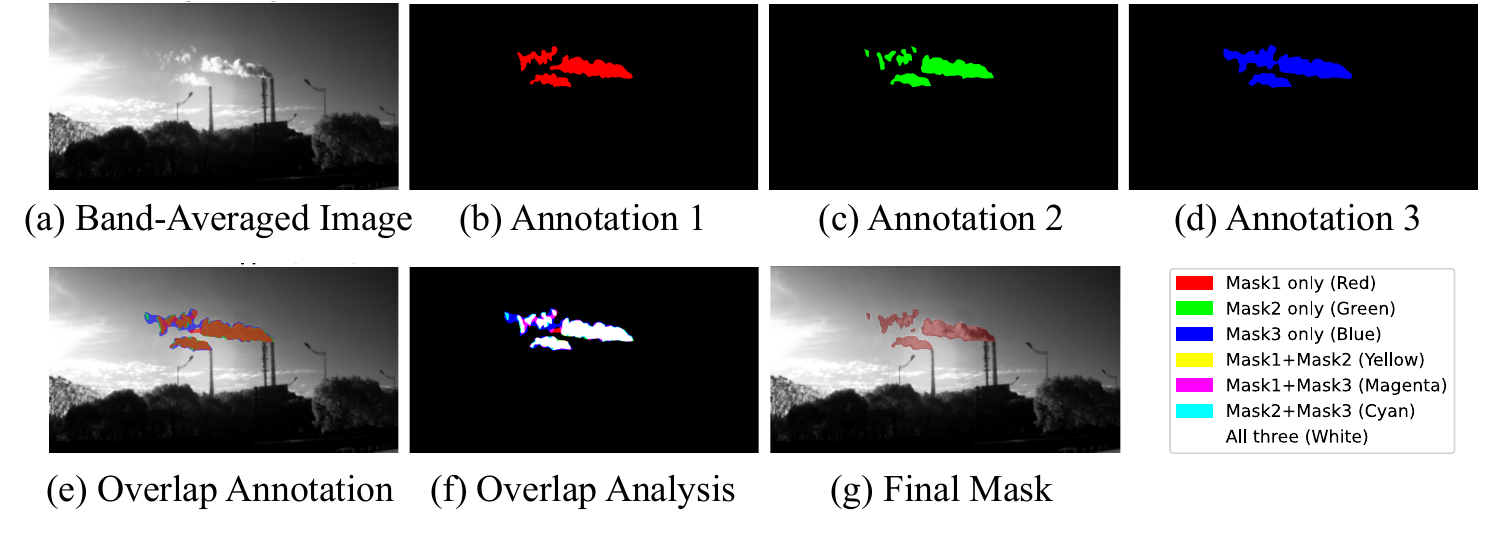}
    \end{center}
    \caption{Many-to-One annotations for hyperspectral smoke segmentation.}
    \label{fig:annotation}
\end{figure}

\noindent\textbf{Many-to-One Annotations.} To ensure annotation quality and handle the inherent ambiguity in smoke boundary delineation, we recruit nine trained expert annotators and implement a protocol where each selected frame receives three independent ground truth masks from three different annotators. All annotations are performed on the band-averaged grayscale images generated from the hyperspectral data. This multi-annotator approach generates three separate segmentation masks per image, with special emphasis placed on challenging regions including early-stage minimal smoke, semi-transparent regions, and blurred boundaries. 
Figure~\ref{fig:annotation} illustrates our Many-to-One Annotations process. The first row shows (a) the band-averaged grayscale image generated by averaging all spectral bands and (b-d) the three independent annotations from different annotators. As observed, different annotators exhibit varying judgments in transparent regions and smoke boundary areas, highlighting the inherent ambiguity in smoke segmentation. The overlapped annotation visualization (e) displays all masks superimposed together, while the overlap analysis (f) uses different colors to represent regions with varying consensus levels, clearly revealing that annotators have significant disagreements in transparent areas and edge regions. The final mask (g) represents our ground truth, where each pixel is classified as smoke if at least two annotators identify it as smoke, otherwise it is classified as background.

\noindent\textbf{Ground Truth Definition.} Final ground truth masks are generated through \textit{majority voting}, where each pixel is classified as smoke if at least two-thirds of annotators label it as smoke. The resulting binary masks assign value 1 to smoke regions and value 0 to background areas. This approach effectively handles annotation uncertainty while preserving reliable boundaries, ultimately producing 1,007 high-quality annotated hyperspectral samples.

\noindent\textbf{Inter-Annotator Agreement Statistics.} Among all pixels marked as smoke by at least one annotator, unanimous agreement (3/3) accounts for 52.07\%, majority agreement (2/3) for 14.14\%, and single annotator only (1/3) for 33.79\%. In terms of the final ground truth (pixels with $\geq$2/3 agreement), 78.63\% come from unanimous agreement and 21.37\% from majority agreement, confirming that our Many-to-One majority voting approach effectively filters uncertain annotations.

\subsection{Multispectral Smoke Segmentation Dataset (MSSDataset)}\label{sec:mssdataset}

\begin{figure}[t]
    \centering
    \includegraphics[width=0.80\linewidth]{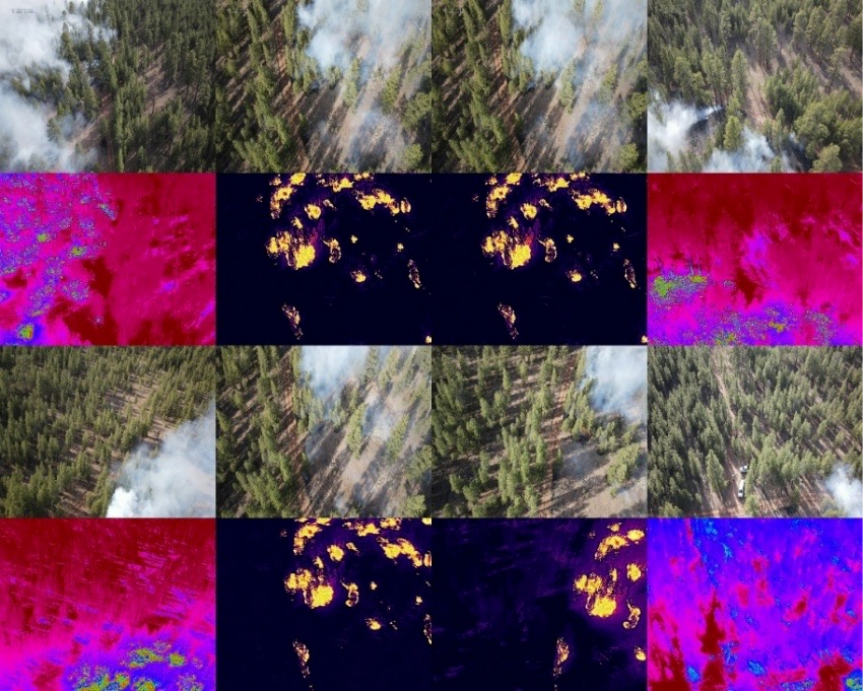}
    \caption{Visible and infrared frame pairs in MSSDataset.}
    \label{fig:mss_dataset}
\end{figure}

To further validate the effectiveness of our proposed method under different spectral configurations, we construct an additional multispectral dataset (MSSDataset) with raw data from FLAME2~\cite{flame2}. Since FLAME2 provides only frame-level classification labels without pixel-level segmentation annotations, we carefully select 200 samples from diverse scenarios and annotate pixel-level smoke masks.
Figure~\ref{fig:mss_dataset} shows representative RGB-IR frame pairs with enhanced IR visualization.
Unlike the 25-band near-infrared configuration of HSSDataset, MSSDataset adopts a 4-channel multispectral configuration combining RGB and infrared (IR) channels, representing a more common multispectral imaging scheme. The infrared channel provides thermal radiation information unavailable in visible-light bands, effectively complementing visible-light perception limitations, especially under nighttime or low-light conditions.
The final MSSDataset provides multispectral cubes encompassing RGB plus infrared channels, allowing us to evaluate our method's performance on different spectral data compositions beyond the 25-band hyperspectral setting.

\section{Methodology}\label{sec4}

In this section, we first present the overall idea of MoP (Sec.~\ref{sec:mop}) and then detail its three key components: band split (BS, Sec.~\ref{sec:band_split}), prototype-based spectral representation (PSR, Sec.~\ref{sec:proto_repr}), and dual-stage router (DSR, Sec.~\ref{sec:dual_router}). Finally, we describe the decoder head and loss function (Sec.~\ref{sec:loss}).

\subsection{Overview of Mixture of Prototypes (MoP)}\label{sec:mop}

\begin{figure*}[t]
    \begin{center}
        \includegraphics[width=\linewidth]{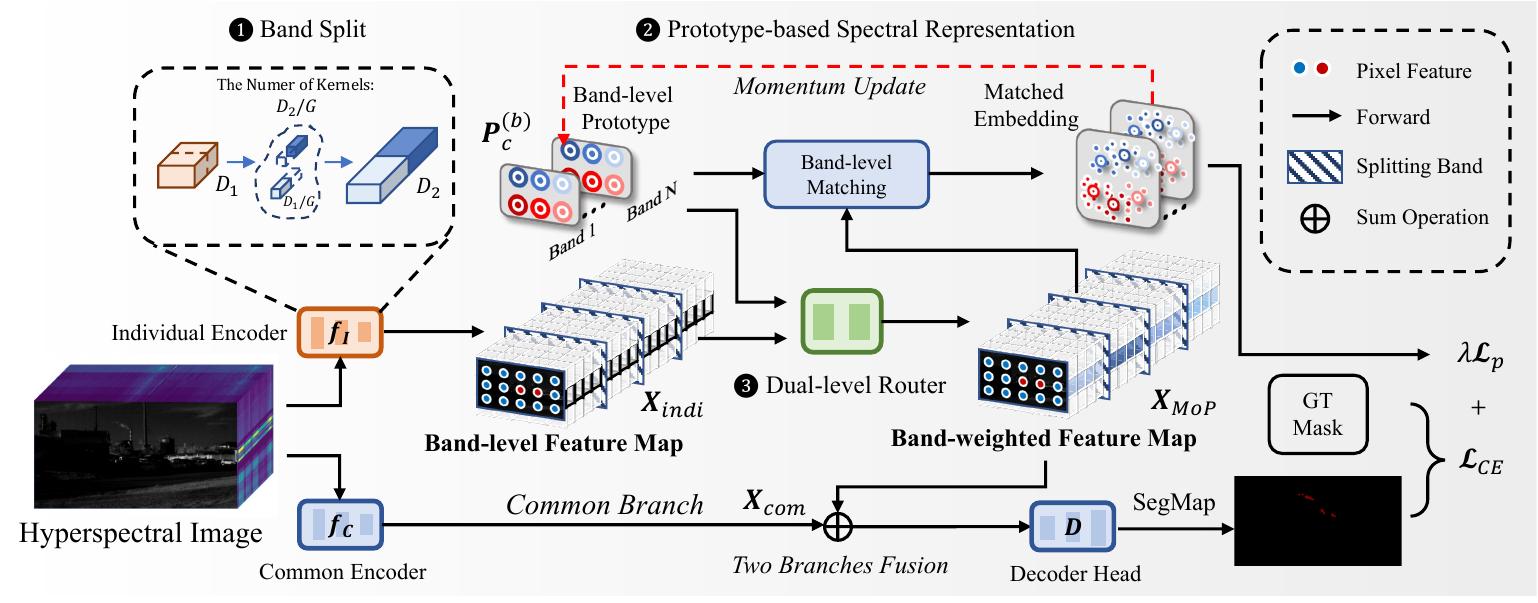}
    \end{center}
    \caption{Overall architecture of our mixture of prototypes (MoP) network.}
    \label{fig:overall_pipeline}
\end{figure*}

The large number of spectral bands in hyperspectral data introduces high-dimensional challenges and potential redundancy, and different bands contribute unequally to specific tasks. Therefore, how to adaptively leverage spectral information is the key challenge for hyperspectral smoke segmentation. Prototype learning captures class distributions by learning representative prototype vectors in feature space, where multi-prototype representations naturally model the highly variable visual patterns and complex intra-class diversity of smoke. Meanwhile, Mixture of Experts (MoE) selectively activates specialized expert networks through gating and routing mechanisms, enabling adaptive specialization for diverse smoke patterns.

Our proposed MoP integrates prototype learning with MoE: prototype learning provides spectral reference representations for hyperspectral data, while the MoE routing mechanism provides adaptive scheduling for these prototypes, enabling the model to dynamically invoke matching spectral patterns based on feature differences across spatial regions.

As illustrated in Figure~\ref{fig:overall_pipeline}, given a hyperspectral image $\bm{I} \in \mathbb{R}^{h \times w \times d}$ ($d=25$), our MoP pipeline operates through the following stages: First, the input is processed by two parallel branches: (1) individual branch: an individual encoder $\bm{f}_{I}$ that applies band split to extract features from each spectral band independently, preserving spectral signatures ($\bm{X}_{\text{indi}}$); and (2) common branch: a common encoder $\bm{f}_{C}$ that follows conventional segmentation approaches, processing all spectral bands jointly to provide general spatial-semantic understanding ($\bm{X}_{\text{com}}$). The common branch adopts a reduced MiT~\cite{xie2021segformer} encoder, with stage-wise channels and encoder block depths reduced to match the individual branch, ensuring consistent feature dimensions and modeling capacity across both branches. Next, features from the individual branch undergo band weighting through our DSR, which takes $\bm{X}_{\text{indi}}$ and band-level prototypes as inputs to generate weighted features $\bm{X}_{\text{MoP}}$. The prototypes are dynamically updated through momentum-based learning during training while remaining fixed during inference. Finally, the weighted features $\bm{X}_{\text{MoP}}$ are fused with features $\bm{X}_{\text{com}}$ from the common branch and passed through a decoder head (a $1 \times 1$ convolution that maps from $D$ dimensions to 2 classes) to produce the final segmentation results.

\subsection{Band Split (BS)}\label{sec:band_split}

\noindent\textbf{Intuition.} Conventional deep learning backbones, including convolution and transformer architectures, inherently perform feature aggregation across the full channel dimension when processing hyperspectral data. Such cross-channel computations inevitably cause signal mixing between different spectral bands, producing ``spectral pollution'' where each spatial location in the output feature map becomes a coupled result of all input band features. This mixing forces the subtle spectral differences between bands to blend, losing the ability to capture band-specific spectral responses. To mitigate spectral pollution, we propose a band split strategy that processes each spectral band independently. 

\noindent\textbf{Implementation.} 

\begin{figure}[t]
    \begin{center}
        \includegraphics[width=0.80\linewidth]{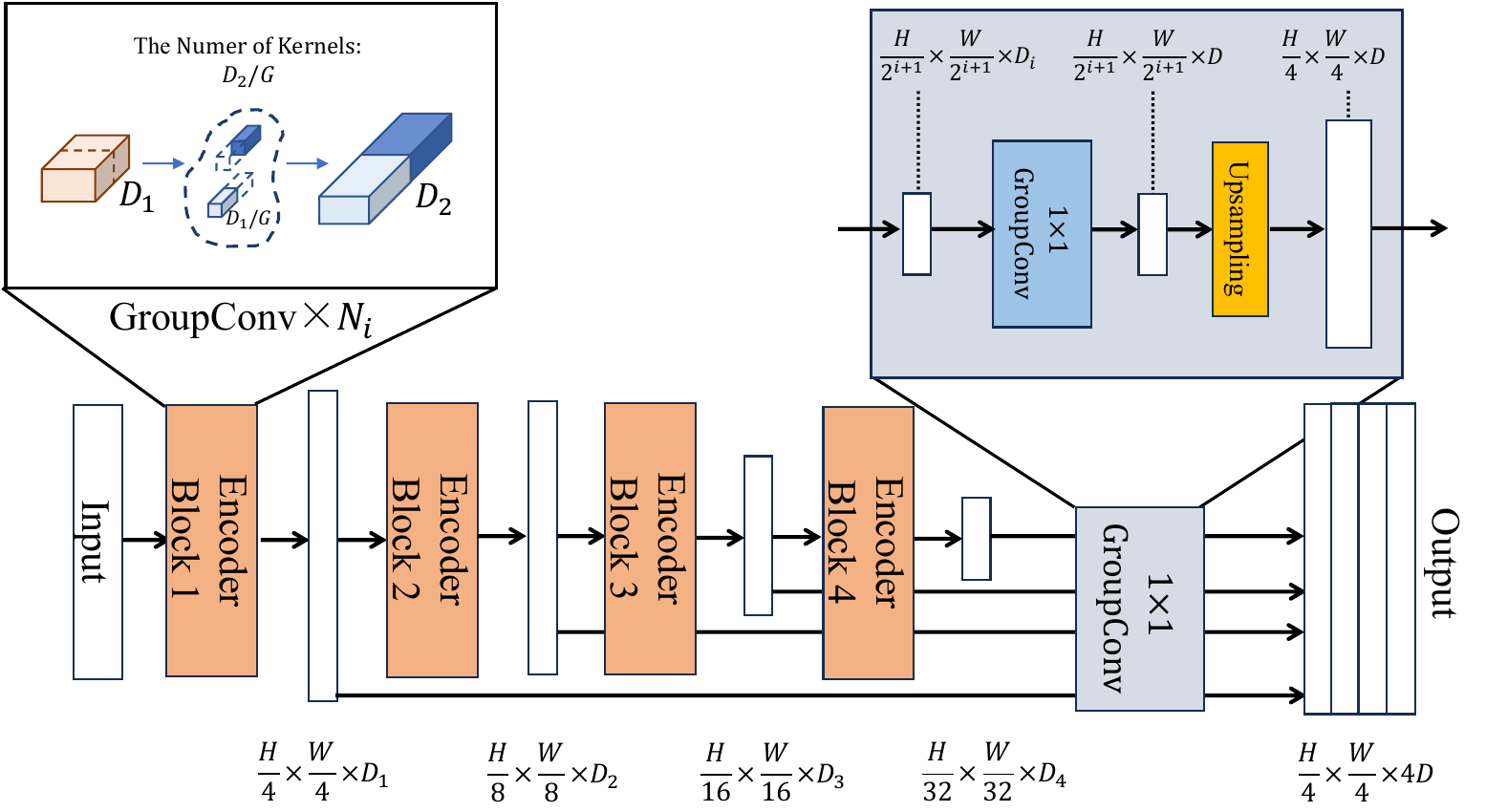}
    \end{center}
    \caption{The individual branch is built upon the MiT architecture with key modifications to achieve band split through group convolution with the number of groups equal to the band count ($G=25$).}
    \label{fig:spectral_branch}
\end{figure}

\noindent\textbf{Individual Branch Design.} The key to achieving band split lies in finding a computational structure that blocks cross-band information flow during feature extraction. Noting that group convolution naturally possesses the property of channel-wise group isolation, when the number of groups $G$ is set strictly equal to the number of input spectral bands $d$, each group of convolution kernels processes only a single band's input signal, thereby achieving complete decoupling of cross-band signals at the structural level while precisely preserving each band's spectral discriminability.

As shown in Figure~\ref{fig:spectral_branch}, the individual branch is built upon the MiT~\cite{xie2021segformer} four-stage architecture, constructing a group convolution encoder. The encoder consists of four cascaded encoding blocks, with spatial downsampling at strides 4, 2, 2, 2 across stages, producing multi-scale feature maps with output channels $D_i$ ($i=1,2,3,4$). Each encoding block contains a downsampling layer and $N_i$ group convolution modules. The downsampling layer performs spatial resolution reduction through strided group convolution with $G=d=25$ groups. Each group convolution module comprises a $1\times1$ channel-expansion group convolution, a $3\times3$ group convolution, and a $1\times1$ channel-reduction group convolution in sequence, with all convolutions using $G=d=25$ groups to ensure channel-level isolation across bands.

We design two configuration variants, real-time and accuracy-oriented, sharing the same stage-wise channel dimensions $\{D_i\}_{i=1}^{4} = \{25, 50, 125, 200\}$, differing only in encoder block depth: the real-time configuration uses $\{N_i\}_{i=1}^{4} = \{2, 2, 2, 2\}$, while the accuracy-oriented configuration uses $\{N_i\}_{i=1}^{4} = \{3, 4, 16, 3\}$. The multi-scale features from the four encoding blocks are each aligned to a unified channel dimension $D=250$ via $1\times1$ group convolution, upsampled to the same spatial resolution, concatenated along the channel dimension, and finally fused through a $1\times1$ group convolution to produce the output features $\bm{X}_{\text{indi}}$.

\noindent\textbf{Feature Representation.} For convenience in subsequent computations, we uniformly adopt the flattened representation, which transforms 2D spatial feature maps into 1D sequences to facilitate pixel-prototype matching via the Sinkhorn algorithm. The individual branch output features $\bm{X}_{\text{indi}} \in \mathbb{R}^{D \times H \times W}$ are reshaped to $\mathbb{R}^{D \times N}$, where $N = H \times W$ is the number of spatial locations. Each spectral band is represented by exactly $D/d$ channels. For band $\ell$, we denote the band-specific features by $\bm{X}_{\text{indi}}^{(\ell)} \in \mathbb{R}^{(D/d) \times N}$. Correspondingly, the feature component of the $n$-th pixel at band $\ell$ is denoted as $\bm{x}_{n}^{(\ell)} \in \mathbb{R}^{D/d}$, and concatenating all band feature components along the channel dimension yields the full spectral feature for that pixel:
\begin{equation}
\bm{x}_{n} = [\bm{x}_{n}^{(1)};\, \bm{x}_{n}^{(2)};\, \ldots;\, \bm{x}_{n}^{(d)}] \in \mathbb{R}^{D}.
\end{equation}

\subsection{Prototype-based Spectral Representation (PSR)}\label{sec:proto_repr}

\noindent\textbf{Intuition.} The spectral response of smoke is simultaneously influenced by multiple physical factors including concentration, particle size distribution, chemical composition, and environmental temperature. The feature distributions across different bands differ significantly in morphology and statistical structure, and a single-center representation cannot adequately capture these diverse spectral patterns. We propose {prototype-based spectral representation}, which learns \textit{band-level prototypes}, a set of representative feature centers for each spectral band. This multi-centric representation not only maintains structural consistency of feature distributions at the global scale but also preserves the discriminative information specific to each band at fine granularity.

\noindent\textbf{Implementation.} 

\noindent\textbf{Prototype Definition.}
Let $\Omega$ denote the number of classes and $\omega \in \{1, 2, \cdots, \Omega\}$ the class index. For each class $\omega$ and each spectral band $\ell$, we define $K$ prototypes forming the prototype matrix $\bm{P}^{\omega,(\ell)} = [\bm{p}_{k}^{\omega,(\ell)}]_{k=1}^{K} \in \mathbb{R}^{(D/d) \times K}$.

\noindent\textbf{Pixel-Prototype Matching.}
For each spectral band, we perform pixel-level matching between spectral features and prototypes to establish correspondence and update prototype representations. Since matching is only performed during the training phase, ground truth labels $\bm{Y}$ are used to group pixel features by class. The pixel feature matrix for class $\omega$ at band $\ell$ is $\bm{X}_{\text{indi}}^{\omega,(\ell)} = [\bm{x}_{n}^{\omega,(\ell)}]_{n=1}^{N^{\omega,(\ell)}} \in \mathbb{R}^{(D/d) \times N^{\omega,(\ell)}}$, where $N^{\omega,(\ell)}$ is the number of pixels belonging to class $\omega$ at band $\ell$.
The matching process employs the Sinkhorn-Knopp algorithm to solve the optimal transport problem~\cite{cuturi2013sinkhorn}:
\begin{equation}\small
\begin{aligned}
\bm{M}^{\omega,(\ell)} = \text{Sinkhorn-Knopp}\left(\bm{P}^{\omega,(\ell)\top} \bm{X}_{\text{indi}}^{\omega,(\ell)}, \varepsilon\right),
\end{aligned}
\end{equation}
where $\bm{P}^{\omega,(\ell)} \in \mathbb{R}^{(D/d) \times K}$ represents the prototype matrix for class $\omega$ and band $\ell$, $\bm{X}_{\text{indi}}^{\omega,(\ell)} \in \mathbb{R}^{(D/d) \times N^{\omega,(\ell)}}$ denotes the spectral feature matrix, and $\bm{M}^{\omega,(\ell)} \in \mathbb{R}^{K \times N^{\omega,(\ell)}}$ is the resulting assignment matrix. During inference, the prototypes remain fixed and are used directly in the DSR without requiring class-specific feature extraction.

\noindent\textbf{Prototype Learning and Update.} The prototypes are initialized using truncated normal initialization with standard deviation 0.02. The prototypes are dynamically updated during training through a momentum-based learning mechanism:
\begin{equation}\small
\begin{aligned}\label{eq:update}
\bm{p}_{k}^{\omega,(\ell)} \leftarrow  \mu\bm{p}_{k}^{\omega,(\ell)} + (1-\mu)\bar{\bm{x}}_{k}^{\omega,(\ell)},
\end{aligned}
\end{equation}
where $\mu=0.999$ is the momentum coefficient, and $\bar{\bm{x}}_{k}^{\omega,(\ell)}$ indicates the $\ell_2$-normalized, mean vector of the embedded training pixels assigned to prototype $\bm{p}_{k}^{\omega,(\ell)}$ by pixel-prototype matching. This momentum update strategy requires no additional backpropagation, smoothly tracking the long-term distribution changes of spectral features across bands while maintaining training stability.

\subsection{Dual-Stage Router (DSR)}\label{sec:dual_router}

\begin{figure}[t]
    \begin{center}
        \includegraphics[width=0.9\linewidth]{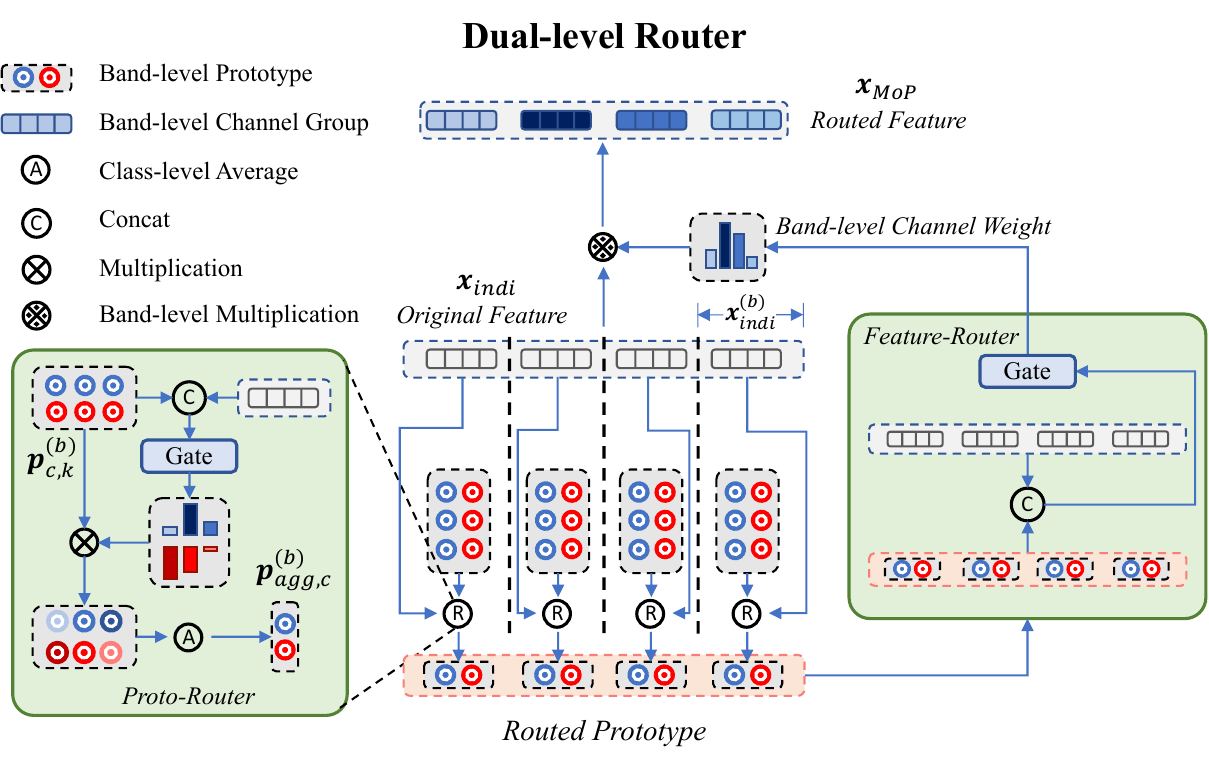}
    \end{center}
    \caption{The pipeline of the dual-stage router (DSR). \textit{Note: For visual clarity, this schematic illustrates the routing process using four spectral bands as a simplified example; the actual system operates on all 25 spectral bands.}}
    \label{fig:mop_pipeline}
\end{figure}

\noindent\textbf{Intuition.} 
Given band-split features and multiple band-level prototypes, how to efficiently integrate both is the core problem. A naive approach of weighting based on pixel-prototype similarities faces the challenge that $d$ bands each containing $K$ prototypes (e.g., $d=25, K=3$ yields 75 prototypes) results in a prohibitively large number of prototypes, and simple similarity computation cannot capture complex cross-band spectral dependencies. Our DSR decomposes this problem into prototype aggregation and band weighting as two sequential sub-tasks.

Note that unlike the pixel-prototype matching in PSR (which requires label-based class grouping), the DSR operates during both training and inference without requiring label information. Therefore, pixel features are represented directly as band features $\bm{x}_{n}^{(\ell)}$ and their full-band concatenation $\bm{x}_{n} = [\bm{x}_{n}^{(1)};\,\ldots;\,\bm{x}_{n}^{(d)}]$ (without class superscript $\omega$), while prototypes retain their class superscript $\omega$ since their class affiliation is fixed from training.

\noindent\textbf{Implementation.} 

\noindent\textbf{Dual-Stage Router Architecture.} As illustrated in Figure~\ref{fig:mop_pipeline}, the DSR consists of two sequential stages: the prototype router (P-Router) and the feature router (F-Router), each addressing distinct optimization challenges.

\noindent\textbf{Stage 1: Prototype Router (P-Router).} The P-Router operates at the band level to adaptively aggregate the $K$ prototypes per class into a single representative prototype for each band. For each spectral band $\ell$, the pixel feature $\bm{x}_{n}^{(\ell)} \in \mathbb{R}^{D/d}$ is concatenated with all prototypes of that band along the feature dimension, and passed through a $1\times 1$ convolution layer followed by Softmax normalization to obtain the weight vector:
\begin{equation}\small
\begin{aligned}
\boldsymbol{\alpha}^{(\ell)} = \text{Softmax}\!\left(\text{Conv}\!\left(\left[\bm{x}_{n}^{(\ell)};\, \bm{p}_{1}^{1,(\ell)};\,\ldots;\,\bm{p}_{K}^{\Omega,(\ell)}\right],\, \theta_{\text{p-router}}\right)\right) \in \mathbb{R}^{\Omega K},
\end{aligned}
\end{equation}
where $\text{Conv}(\cdot,\theta)$ denotes a convolution operation with parameters $\theta$, and $\theta_{\text{p-router}}$ represents the parameters of a $1\times 1$ convolution with $(\Omega K+1)(D/d)$ input channels and $\Omega K$ output channels. The scalar $\alpha_{k}^{\omega,(\ell)}$ represents the aggregation weight of the $k$-th prototype of class $\omega$ at band $\ell$, satisfying $\sum_{\omega=1}^{\Omega}\sum_{k=1}^{K}\alpha_{k}^{\omega,(\ell)}=1$. The aggregated prototype for each class $\omega$ at band $\ell$ is:
\begin{equation}\small
\begin{aligned}
\bm{p}_{\text{agg}}^{\omega,(\ell)} = \sum_{k=1}^{K} \alpha_{k}^{\omega,(\ell)} \cdot \bm{p}_{k}^{\omega,(\ell)}.
\end{aligned}
\end{equation}
The aggregated prototypes across all bands are concatenated to form the complete aggregated prototype $\bm{p}_{\text{agg}}^{\omega} = [\bm{p}_{\text{agg}}^{\omega,(1)};\,\ldots;\,\bm{p}_{\text{agg}}^{\omega,(d)}] \in \mathbb{R}^{D}$.

\noindent\textbf{Stage 2: Feature Router (F-Router).} The F-Router operates on the complete feature representation to generate adaptive band-level weighting. The full-band concatenated feature $\bm{x}_{n} \in \mathbb{R}^{D}$ is concatenated with the aggregated prototypes of all classes $\bm{p}_{\text{concat}} = [\bm{p}_{\text{agg}}^{1};\,\ldots;\,\bm{p}_{\text{agg}}^{\Omega}] \in \mathbb{R}^{\Omega D}$, and passed through a $1\times 1$ convolution layer with Softmax normalization to output band-level importance weights:
\begin{equation}\small
\begin{aligned}
\boldsymbol{\beta} = \text{Softmax}\!\left(\text{Conv}\!\left(\left[\bm{x}_{n};\, \bm{p}_{\text{concat}}\right],\, \theta_{\text{f-router}}\right)\right) \in \mathbb{R}^{d},
\end{aligned}
\end{equation}
where $\theta_{\text{f-router}}$ represents the parameters of a $1\times 1$ convolution with $(1+\Omega)D$ input channels and $d$ output channels. The $\ell$-th component $\beta^{(\ell)}$ corresponds to the importance weight of the $\ell$-th spectral band, satisfying $\sum_{\ell=1}^{d}\beta^{(\ell)} = 1$.

\noindent\textbf{Final Output.} The DSR produces adaptively weighted spectral features through group multiplication, applying each band weight $\beta^{(\ell)}$ to the corresponding band's feature group:
\begin{equation}\small
\begin{aligned}
\tilde{\bm{x}}_{n} = [\beta^{(1)} \cdot \bm{x}_{n}^{(1)};\, \beta^{(2)} \cdot \bm{x}_{n}^{(2)};\, \ldots;\, \beta^{(d)} \cdot \bm{x}_{n}^{(d)}] \in \mathbb{R}^{D}.
\end{aligned}
\end{equation}
Note that the above description uses a single pixel feature as an example for clarity. In practice, we extend this operation to the entire image level.

\subsection{Decoder Head and Loss Function}\label{sec:loss}

\noindent\textbf{Dual-Branch Feature Fusion.} 
Our decoder head processes features from two parallel branches to generate the final segmentation predictions. The architecture integrates both common spectral features $\bm{X}_{\text{com}}$ and MoP-enhanced spectral features $\bm{X}_{\text{MoP}} = [\tilde{\bm{x}}_{n}]_{n=1}^{N} \in \mathbb{R}^{D \times N}$ through element-wise averaging:
\begin{equation}\small
\bm{X}_{\text{fusion}} = \frac{\bm{X}_{\text{com}} + \bm{X}_{\text{MoP}}}{2}
\end{equation}
The fused features $\bm{X}_{\text{fusion}}$ are reshaped to spatial feature maps and processed through a simple but effective decoder head consisting of a convolutional layer followed by softmax activation to generate the final segmentation predictions.

\noindent\textbf{Loss Function.} 
Our training objective consists of two complementary components:

\noindent\textbf{Pixel-Prototype Contrastive Learning.} The pixel-prototype contrastive loss enforces discriminative feature space by pulling each pixel feature $\bm{x}_{n,k}^{\omega,(\ell)}$ close to its assigned prototype $\bm{p}_{k}^{\omega,(\ell)}$ and pushing it away from other negative prototypes $\mathcal{P}^{-}$:
\begin{equation}\small
    \begin{aligned}\label{eq:ppc}
    \mathcal{L}_{\text{P}} = -\frac{1}{N \times d}\sum_{\omega=1}^{\Omega}\sum_{k=1}^{K}\sum_{\ell=1}^{d}\sum_{n=1}^{N_{k}^{\omega,(\ell)}}\log \frac{\exp(\bm{x}_{n,k}^{\omega,(\ell)\top}\bm{p}_{k}^{\omega,(\ell)}/\tau)}{\exp(\bm{x}_{n,k}^{\omega,(\ell)\top}\bm{p}_{k}^{\omega,(\ell)}/\tau) + \sum_{\bm{p}^{-}\in\mathcal{P}^{-}} \exp(\bm{x}_{n,k}^{\omega,(\ell)\top}\bm{p}^{-}/\tau)},
    \end{aligned}
    \end{equation}%
where $\bm{x}_{n,k}^{\omega,(\ell)}$ denotes the individual branch feature of the $n$-th pixel matched to the $k$-th prototype for class $\omega$ at band $\ell$, and $\mathcal{P}^{-}$ is the set of all prototypes at the current band except the matched one. Both features and prototypes are $\ell_2$-normalized, and the temperature $\tau=0.1$ controls the concentration level. Since the denominator sums over the matched prototype and all prototypes in $\mathcal{P}^{-}$, covering all $\Omega K$ prototypes at the current band, Eq.~(\ref{eq:ppc}) can be equivalently expressed as a standard cross-entropy form for efficient computation:
\begin{equation}\small\label{eq:ppc_ce}
\mathcal{L}_{\text{P}} = \frac{1}{N \times d}\sum_{\ell=1}^{d}\sum_{n=1}^{N} \operatorname{CrossEntropy}\!\left(\left[\bm{x}_{n}^{(\ell)\top}\bm{p}_{k}^{\omega,(\ell)}/\tau\right]_{\forall\,\omega,k},\; y_{n}^{(\ell)}\right),
\end{equation}
where $y_{n}^{(\ell)}$ is the prototype index assigned by Sinkhorn matching.

\noindent\textbf{Segmentation Loss.} For the final segmentation task, we employ the standard binary cross-entropy loss:
\begin{align}
\mathcal{L}_{\text{BCE}} = -\frac{1}{N} \sum_{n=1}^{N} \left[ y_n \log(\hat{y}_n) + (1-y_n) \log(1-\hat{y}_n) \right],
\end{align}%
where $\hat{y}_n$ represents the predicted probability for pixel $n$, and $y_n$ is the ground truth label.

\noindent\textbf{Total Loss.} The complete training objective combines both components:
\begin{align}
\mathcal{L}_{\text{total}} = \mathcal{L}_{\text{BCE}} + \lambda \mathcal{L}_{\text{P}},
\end{align}%
where $\lambda$ is set to 0.01 following the configuration established in prior prototype learning literature~\cite{zhou2024prototype}. 

\section{Experiments}\label{sec5}

\noindent This section provides experiments and analysis. Section~\ref{sec:exp_setup} presents the datasets, baselines, implementation, and backbone configurations. Section~\ref{sec:benchmark} reports overall benchmark results on HSSDataset and MSSDataset, including comparisons under real-time vs accuracy-oriented settings, scale-aware analysis, and qualitative visualizations. Section~\ref{sec:ablation} conducts module-wise ablations of BS, PSR, and DSR. Section~\ref{sec:other_ablation} further studies design choices and hyperparameters.

\subsection{Experimental Setup}\label{sec:exp_setup}

\noindent\textbf{Datasets.} Our experiments are conducted on two datasets: (1) {HSSDataset}: Our newly constructed hyperspectral smoke segmentation dataset containing 1,007 samples across 25 spectral bands (600-974nm). The dataset covers diverse challenging scenarios, including high exposure, low visibility, early-stage minimal smoke, cloud interference, and complex backgrounds. The annotated samples are split into training, validation, and test sets with a ratio of approximately 6:2:2. (2) {MSSDataset}: A multispectral dataset with RGB-IR paired samples derived from the FLAME2 dataset~\cite{flame2}, providing 4-channel spectral information. Due to the scale of this dataset, the annotated samples are split into training, validation, and test sets with a ratio of approximately 8:1:1 to ensure sufficient training data. To comprehensively evaluate performance, we further conduct scale-aware evaluation by categorizing smoke into small, medium, and large scales, with the partition criteria following FoSp~\cite{yao2024fosp}.

\noindent\textbf{Baseline Comparisons.} As this is the first work for hyperspectral smoke segmentation, we evaluate against various strong segmentation baselines, including CNN-based approaches (PSPNet~\cite{zhao2017pyramid}, DeepLabV3+~\cite{chen2018encoder}, OCRNet~\cite{yuan2020object}, SegNeXt~\cite{guo2022segnext}, Trans-BVM~\cite{yan2022transmission}), Transformer-based methods (SegFormer~\cite{xie2021segformer}, Mask2Former~\cite{cheng2022masked}, ProtoSeg~\cite{zhou2024prototype}, FoSp~\cite{yao2024fosp}), where Trans-BVM and FoSp are specifically designed for smoke segmentation. We reproduce these methods and adapt them to hyperspectral data by only modifying the input channels from 3 to 25 while keeping other components unchanged.

\begin{table*}[t]
    \centering
    \caption{Quantitative comparison on hyperspectral smoke segmentation benchmarks. We compare methods from various domains, including CNN-based and Transformer-based approaches across real-time and accuracy-oriented settings. The best: \textbf{bold}, the second: \underline{underline}.}
    \setlength{\tabcolsep}{3pt}
    \resizebox{\linewidth}{!}{
    \begin{tabular}{l|l|c|cc|cc|cc|cc}
    \toprule[1pt]  
                                & \multirow{2}{*}{Methods} & \multirow{2}{*}{\shortstack{Backbone\\Category}} & \multicolumn{2}{c}{Small}       & \multicolumn{2}{c}{Medium}       & \multicolumn{2}{c}{Large}      & \multicolumn{2}{c}{Total} \\ \cmidrule{4-11}
                                &                          &                           & $F_1$            & $mIoU$           & $F_1$            & $mIoU$           & $F_1$            & $mIoU$           & $F_1$            & $mIoU$ \\ \midrule
\multirow{9}{*}{\textit{Real-Time}}      & PSPNet~\cite{zhao2017pyramid} & \multirow{4}{*}{CNN-based} & 49.32 & 35.87 & 64.14 & 47.80 & 71.39 & 55.93 & 61.23 & 46.07 \\
                                & Deeplabv3+~\cite{chen2018encoder} &  & 53.50 & 39.68 & 63.79 & 47.59 & \underline{73.12} & \underline{58.10} & 62.92 & 47.82 \\
                                & OCRNet~\cite{yuan2020object} &  & 43.83 & 32.17 & 61.23 & 46.62 & 69.47 & 54.52 & 59.26 & 46.20 \\
                                & SegNeXt~\cite{guo2022segnext} &  & 48.24 & 33.17 & 64.74 & 48.38 & 70.57 & 54.86 & 60.80 & 45.06 \\ \cmidrule{2-11}
                                & SegFormer~\cite{xie2021segformer} & \multirow{5}{*}{\shortstack{Transformer-\\based}} & 53.48 & 37.78 & 65.63 & 49.21 & 70.17 & 54.33 & 62.87 & 46.84 \\
                                & FoSp~\cite{yao2024fosp} &  & \underline{56.90} & \underline{40.40} & 65.77 & 49.48 & 72.69 & 55.65 & 64.99 & 48.97 \\
                                & ProtoSeg~\cite{zhou2024prototype} &  & 55.86 & 39.56 & \underline{67.83} & \underline{51.72} & 72.49 & 55.37 & \underline{65.43} & \underline{49.53} \\
                                & \cellcolor{Gray}MoP (ours) & \cellcolor{Gray} & \cellcolor{Gray}\textbf{59.41} & \cellcolor{Gray}\textbf{46.32} & \cellcolor{Gray}\textbf{71.19} & \cellcolor{Gray}\textbf{56.27} & \cellcolor{Gray}\textbf{74.17} & \cellcolor{Gray}\textbf{59.33} & \cellcolor{Gray}\textbf{68.13} & \cellcolor{Gray}\textbf{53.84} \\ \midrule
\multirow{11}{*}{\shortstack{\textit{Accuracy-}\\\textit{Oriented}}} & PSPNet~\cite{zhao2017pyramid} & \multirow{5}{*}{CNN-based} & 59.09 & 43.35 & 70.95 & 55.64 & 77.40 & 63.53 & 68.80 & 53.73 \\
                                & Deeplabv3+~\cite{chen2018encoder} &  & 57.78 & 41.56 & \underline{72.04} & \underline{56.91} & {77.50} & \underline{64.91} & 69.11 & 54.03 \\
                                & Trans-BVM~\cite{yan2022transmission} &  & 58.76 & 43.51 & 71.81 & 56.79 & 76.41 & 64.66 & 69.44 & 54.80 \\
                                & OCRNet~\cite{yuan2020object} &  & 48.27 & 37.58 & 68.27 & 53.34 & 77.51 & 64.15 & 64.20 & 51.08 \\
                                & SegNeXt~\cite{guo2022segnext} &  & 62.54 & 47.04 & 68.59 & 53.43 & 73.92 & 59.37 & 68.72 & 53.57 \\ \cmidrule{2-11}
                                & Mask2Former~\cite{cheng2022masked} & \multirow{6}{*}{\shortstack{Transformer-\\based}} & 57.31 & 41.47 & 69.41 & 53.63 & 76.44 & 62.16 & 67.34 & 51.94 \\
                                & SegFormer~\cite{xie2021segformer} &  & 64.99 & 48.95 & 67.80 & 51.68 & 75.34 & 60.75 & 68.90 & 53.21 \\
                                & FoSp~\cite{yao2024fosp} &  & \underline{67.34} & \underline{51.72} & 70.60 & 55.01 & 77.48 & 63.50 & \underline{71.38} & \underline{56.21} \\
                                & ProtoSeg~\cite{zhou2024prototype} &  & 66.00 & 51.27 & 68.96 & 53.18 & \underline{77.54} & 63.48 & 71.15 & 56.11 \\
                                & \cellcolor{Gray}MoP (ours) & \cellcolor{Gray} & \cellcolor{Gray}\textbf{69.32} & \cellcolor{Gray}\textbf{54.56} & \cellcolor{Gray}\textbf{73.89} & \cellcolor{Gray}\textbf{59.03} & \cellcolor{Gray}\textbf{78.50} & \cellcolor{Gray}\textbf{65.15} & \cellcolor{Gray}\textbf{73.63} & \cellcolor{Gray}\textbf{59.21} 
    \\ \bottomrule[1pt]
    \end{tabular}}
    \label{tab:HSS_main_results}
\end{table*}

\noindent\textbf{Implementation Details.}
We implement our method using PyTorch on the MMSegmentation. All experiments are conducted on NVIDIA RTX 3090Ti GPUs. For data augmentation during training, we adopt standard techniques including random horizontal flipping, random scaling (0.5-2.0), and random cropping. The input hyperspectral images are resized to $512 \times 512$ for training, while maintaining the original aspect ratio during inference.
We utilize the AdamW optimizer with an initial learning rate of 6e-5, weight decay of 0.01, and momentum parameters $\beta_1 = 0.9$ and $\beta_2 = 0.999$. The learning rate is scheduled using polynomial annealing with a power of 0.9. Training is conducted for 40,000 iterations with a batch size of 4. For the MoP, we set the number of prototypes $K=3$ and the feature dimension $D=250$.

\noindent\textbf{Backbone configuration.} As shown in Table~\ref{tab:HSS_main_results}, to ensure fair comparison, we organize methods into two categories: real-time and accuracy-oriented settings, with backbone-specific comparisons within each category. For the real-time setting, we adopt a customized MiT-B0~\cite{xie2021segformer}, while for the accuracy-oriented setting, we use a customized MiT-B3. Both settings modify the patch-embedding to accept 25-channel hyperspectral input and use stage-wise channels \texttt{in\_channels} = [25, 50, 125, 200]. Other architectural components remain unchanged, and the individual branch follows the depths defined in Sec.~\ref{sec:band_split}.

\begin{figure*}[t]
    \begin{center}
        \includegraphics[width=\linewidth]{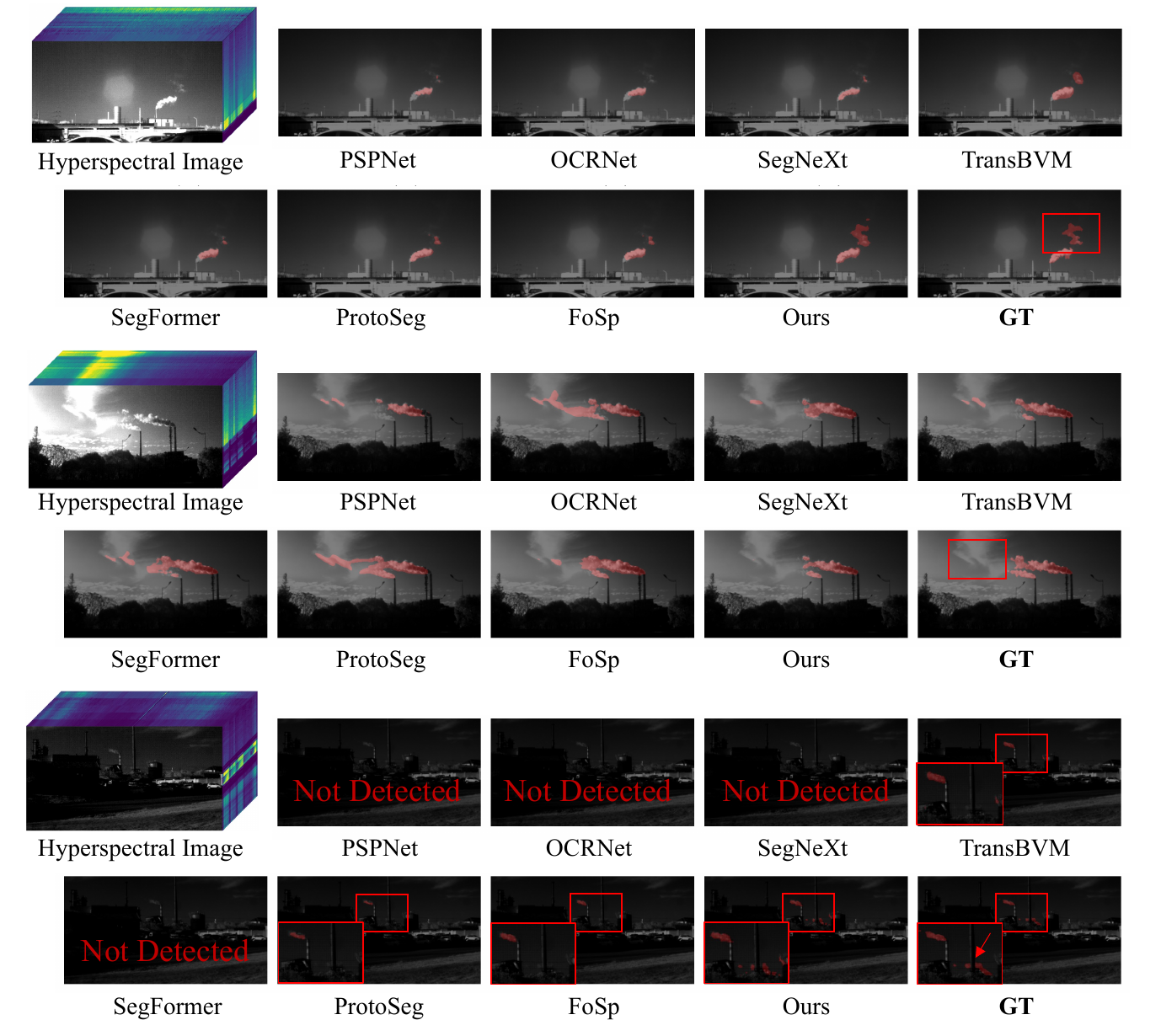}
    \end{center}
    \caption{Visualization of HSSDataset segmentation results. For better visualization, the segmentation masks are overlaid on the band-averaged images.}
    \label{fig:hss_seg}
\end{figure*}

\noindent\textbf{Evaluation Metrics.}
Following previous literature~\cite{yan2022transmission,yao2024fosp}, we employ $F_1$ score and $mIoU$ as our primary evaluation metrics.

\subsection{Benchmark Results}\label{sec:benchmark}

\begin{table*}[t]
    \centering
    \caption{Quantitative comparison on MSSDataset. We compare methods from various domains, including CNN-based and Transformer-based approaches across real-time and accuracy-oriented settings. The best: \textbf{bold}, the second: \underline{underline}.}
    \resizebox{\linewidth}{!}{
    \begin{tabular}{l|l|c|cc|cc|cc}
    \toprule[1pt]  
                                & \multirow{2}{*}{Methods} & \multirow{2}{*}{\shortstack{Backbone\\Category}} & \multicolumn{2}{c}{Medium}       & \multicolumn{2}{c}{Large}      & \multicolumn{2}{c}{Total} \\ \cmidrule{4-9}
                                &                          &                           & $F_1$            & $mIoU$           & $F_1$            & $mIoU$           & $F_1$            & $mIoU$ \\ \midrule
\multirow{9}{*}{\textit{Real-Time}}      & PSPNet~\cite{zhao2017pyramid} & \multirow{4}{*}{CNN-based} & 83.56 & \underline{74.44} & 92.02 & 85.82 & 80.67 & 71.92 \\
                                & Deeplabv3+~\cite{chen2018encoder} &  & 80.15 & 67.42 & 89.89 & 81.77 & 81.87 & 70.06 \\
                                & OCRNet~\cite{yuan2020object} &  & 84.09 & 73.96 & 87.97 & 79.23 & \underline{82.80} & 72.12 \\
                                & SegNeXt~\cite{guo2022segnext} &  & 80.11 & 67.27 & 90.15 & 82.16 & 81.34 & 69.63 \\ \cmidrule{2-9}
                                & SegFormer~\cite{xie2021segformer} & \multirow{5}{*}{\shortstack{Transformer-\\based}} & 83.37 & 74.05 & \underline{92.06} & \underline{86.36} & 82.67 & \underline{72.87} \\
                                & FoSp~\cite{yao2024fosp} &  & \underline{84.27} & {74.08} & 90.38 & 82.80 & 82.43 & 71.71 \\
                                & ProtoSeg~\cite{zhou2024prototype} &  & 81.29 & 68.90 & 91.05 & 83.65 & 82.27 & 70.96 \\
                                & \cellcolor{Gray}MoP (ours) & \cellcolor{Gray} & \cellcolor{Gray}\textbf{85.50} & \cellcolor{Gray}\textbf{75.70} & \cellcolor{Gray}\textbf{93.24} & \cellcolor{Gray}\textbf{87.70} & \cellcolor{Gray}\textbf{83.90} & \cellcolor{Gray}\textbf{74.45} \\ \midrule
\multirow{11}{*}{\shortstack{\textit{Accuracy-}\\\textit{Oriented}}} & PSPNet~\cite{zhao2017pyramid} & \multirow{5}{*}{CNN-based} & 83.66 & 73.92 & \underline{93.11} & \underline{87.26} & 85.51 & \underline{76.91} \\
                                & Deeplabv3+~\cite{chen2018encoder} &  & 83.90 & 73.94 & 92.43 & 86.09 & 83.72 & 73.85 \\
                                & SegNeXt~\cite{guo2022segnext} &  & 84.46 & 73.44 & 92.29 & 85.78 & 86.26 & 76.36 \\
                                & OCRNet~\cite{yuan2020object} &  & 83.40 & 71.96 & 91.13 & 83.80 & 85.40 & 75.06 \\
                                & Trans-BVM~\cite{yan2022transmission} &  & 84.01 & 73.30 & 91.78 & 84.88 & 86.19 & 76.26 \\ \cmidrule{2-9}
                                & Mask2Former~\cite{cheng2022masked} & \multirow{6}{*}{\shortstack{Transformer-\\based}} & 81.04 & 68.72 & 90.33 & 82.52 & 84.47 & 73.74 \\
                                & SegFormer~\cite{xie2021segformer} &  & 84.70 & 73.83 & 91.52 & 84.44 & 86.04 & 76.00 \\
                                & FoSp~\cite{yao2024fosp} &  & \underline{84.88} & \underline{74.14} & 91.45 & 84.34 & \underline{86.61} & 76.81 \\
                                & ProtoSeg~\cite{zhou2024prototype} &  & 82.34 & 70.52 & 90.78 & 83.25 & 85.48 & 75.14 \\
                                & \cellcolor{Gray}MoP (ours) & \cellcolor{Gray} & \cellcolor{Gray}\textbf{85.16} & \cellcolor{Gray}\textbf{75.12} & \cellcolor{Gray}\textbf{93.41} & \cellcolor{Gray}\textbf{87.74} & \cellcolor{Gray}\textbf{88.11} & \cellcolor{Gray}\textbf{79.55} 
    \\ \bottomrule[1pt]
    \end{tabular}}
    \label{tab:FLAME_main_results}
\end{table*}

\noindent\textbf{Hyperspectral Smoke Segmentation Dataset.}
As shown in Table~\ref{tab:HSS_main_results}, we evaluate all methods with scale-aware analysis (small, medium, large) to comprehensively assess the ability to handle different smoke scales, especially early-stage small smoke.

In the \textit{real-time} setting, Transformer-based methods generally outperform CNN-based methods, with ProtoSeg achieving the best baseline total performance due to its prototype learning mechanism. Our MoP achieves the best performance across all scales: 59.41\% $F_1$ and 46.32\% mIoU for small smoke, outperforming the previous best ProtoSeg by 2.70\% $F_1$ and 4.31\% mIoU overall (68.13\% vs 65.43\%). 

For \textit{accuracy-oriented} settings, FoSp and ProtoSeg achieve the best and second-best baseline performance respectively, indicating that methods designed for smoke segmentation or prototype learning are more effective. Our MoP achieves 73.63\% $F_1$ and 59.21\% mIoU, outperforming FoSp by 2.25\% and 3.00\% respectively. Particularly noteworthy is the significant improvement in small smoke (69.32\% $F_1$, +1.98\% over FoSp), demonstrating that the adaptive band weighting mechanism is more effective for challenging small-scale targets. All methods show notably lower performance on small-scale smoke compared to medium and large smoke, confirming that early-stage smoke identification remains a highly challenging problem.

\noindent\textbf{Multispectral Smoke Segmentation Dataset.} 
Table~\ref{tab:FLAME_main_results} demonstrates the effectiveness of our method on the MSSDataset, which consists of RGB-IR paired multispectral data. Since the smoke scale in MSS is generally larger, we report results on the medium and large subsets.
Compared to HSSDataset, all methods achieve higher overall performance on MSSDataset due to generally larger smoke scales and more pronounced wildfire smoke features. The performance rankings also shift across datasets, reflecting different methods' sensitivity to data characteristics. Our MoP also demonstrates good generalization on MSSDataset: in the real-time setting, it achieves 83.90\% $F_1$ and 74.45\% mIoU; for accuracy-oriented settings, it achieves 88.11\% $F_1$ and 79.55\% mIoU, outperforming FoSp by 1.50\% in $F_1$ and PSPNet by 2.64\% in mIoU. Even with only four channels (RGB + IR), the strong performance indicates that the infrared channel provides critical complementary information.

\begin{figure*}[t]
    \begin{center}
        \includegraphics[width=\linewidth]{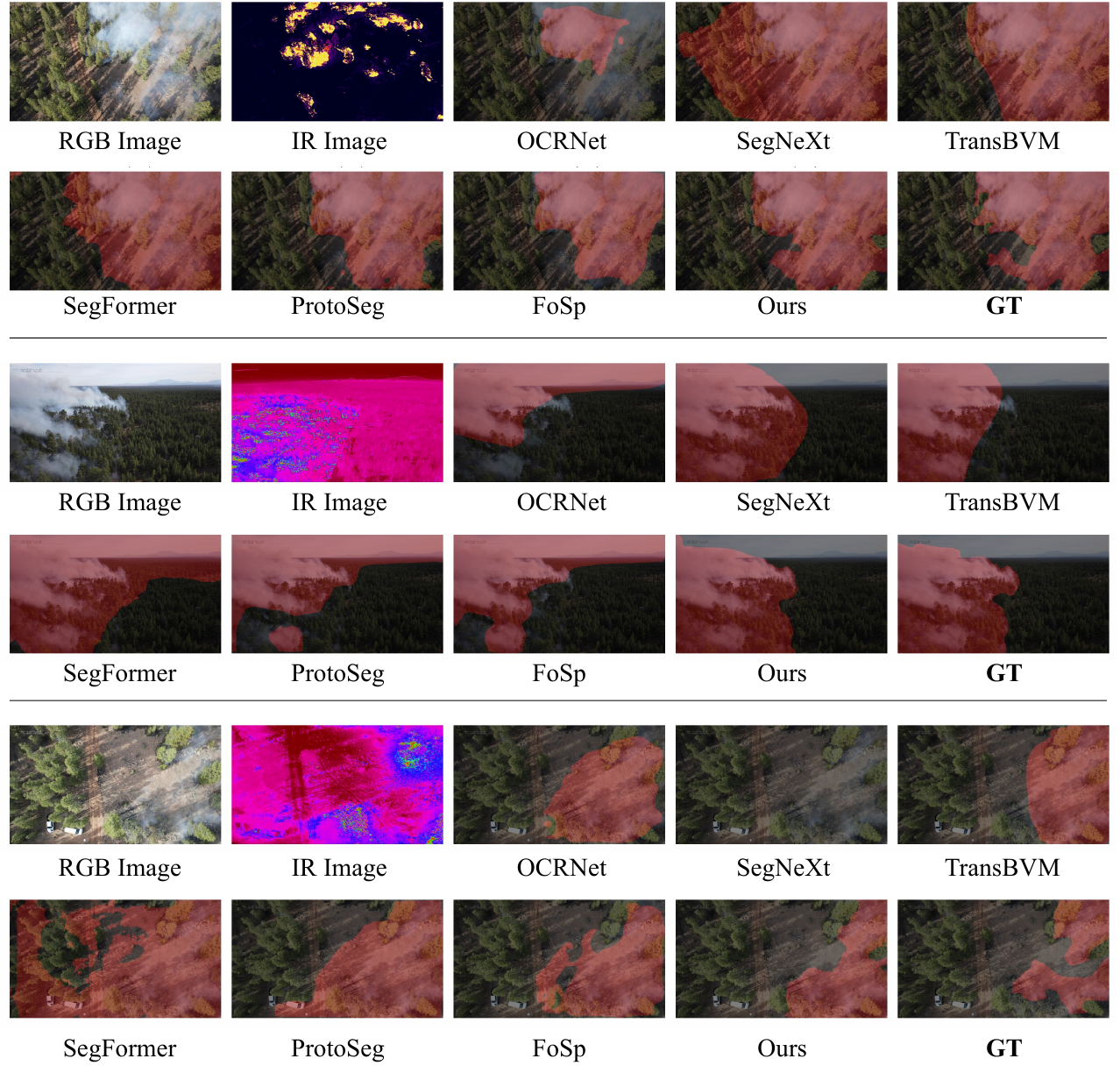}
    \end{center}
    \caption{Visualization of MSSDataset segmentation results. For better visualization, the segmentation masks are overlaid on the RGB images.}
    \label{fig:flame_seg}
\end{figure*}

\noindent\textbf{Qualitative Analysis.} Figure~\ref{fig:hss_seg} presents visual comparisons of our hyperspectral smoke segmentation results on three challenging scenarios in the HSSDataset: transparent smoke, cloud interference, and small smoke in low-light conditions.

For transparent smoke cases, most baseline methods can only detect the main body of smoke but struggle to capture the semi-transparent edge regions. The low contrast of semi-transparent regions against the background makes them difficult to distinguish even in hyperspectral data. In contrast, our MoP, through adaptive band weighting, better handles these challenging regions and successfully captures more semi-transparent edges.

In cloud interference scenarios, many methods tend to misclassify clouds as smoke, producing substantial false positives. Although hyperspectral data provides more spectral information, existing methods do not fully exploit this information to distinguish clouds from smoke. Our MoP, through BS and adaptive weighting, more accurately leverages the discriminative information from different bands, significantly reducing false positives.

For small smoke in low-light conditions, some baseline methods fail to detect the smoke entirely. Even better-performing methods often detect only partial smoke regions. Our MoP shows the best performance on small-scale smoke, benefiting from the DSR's ability to dynamically adjust band weights at different spatial locations, enhancing discriminative features for small targets.

Figure~\ref{fig:flame_seg} shows our method maintains consistent performance on the MSSDataset. Since the smoke scale in MSSDataset is generally larger with more prominent features, most methods can capture the smoke, but MoP still demonstrates advantages in boundary delineation and detail preservation.

\subsection{Ablation Study}\label{sec:ablation}

\begin{table}[t]
    \centering
    \caption{Ablation study of our MoP components on HSS and MSS datasets. The ``bl'' (baseline) refers to the common branch only. ``BS'', ``PSR'', and ``DSR'' denote band split, prototype-based spectral representation, and dual-stage router, respectively; the DSR comprises the feature router (``F-Router'') and the prototype router (``P-Router'').}
    \label{tab:ablation}
    \setlength{\tabcolsep}{4pt}
    \resizebox{\linewidth}{!}{
    \begin{tabular}[t]{c|cccc|cc|cc}
    \toprule[1pt]
    & \multirow{2}{*}{BS} & \multirow{2}{*}{PSR} & \multirow{2}{*}{F-Router} & \multirow{2}{*}{P-Router} & \multicolumn{2}{c|}{HSSDataset} & \multicolumn{2}{c}{MSSDataset} \\
    \cmidrule{6-9}
    & & & & & $F_1$ & $mIoU$ & $F_1$ & $mIoU$ \\ \midrule
    bl & & & & & 68.29 & 53.16 & 84.23 & 73.15 \\
    (a) & \checkmark & & & & \reshl{69.98}{1.69} & \reshl{54.71}{1.55} & \reshl{85.41}{1.18} & \reshl{74.28}{1.13} \\
    (b) & \checkmark & \checkmark & & & \reshl{71.11}{2.82} & \reshl{56.10}{2.94} & \reshl{85.67}{1.44} & \reshl{74.89}{1.74} \\
    (c) & \checkmark & & \checkmark & & \reshl{71.56}{3.27} & \reshl{56.61}{3.45} & \reshl{86.18}{1.95} & \reshl{75.44}{2.29} \\
    (d) & \checkmark & \checkmark & \checkmark & & \reshl{72.91}{4.62} & \reshl{58.46}{5.30} & \reshl{87.42}{3.19} & \reshl{78.41}{5.26} \\
    \cellcolor{Gray}(e) & \cellcolor{Gray}\checkmark & \cellcolor{Gray}\checkmark & \cellcolor{Gray}\checkmark & \cellcolor{Gray}\checkmark & \cellcolor{Gray}\textbf{\reshl{73.63}{5.34}} & \cellcolor{Gray}\textbf{\reshl{59.21}{6.05}} & \cellcolor{Gray}\textbf{\reshl{88.11}{3.88}} & \cellcolor{Gray}\textbf{\reshl{79.55}{6.40}} \\
    \bottomrule[1pt]
    \end{tabular}}
\end{table}

\begin{figure}[t]
    \begin{center}
        \includegraphics[width=\linewidth]{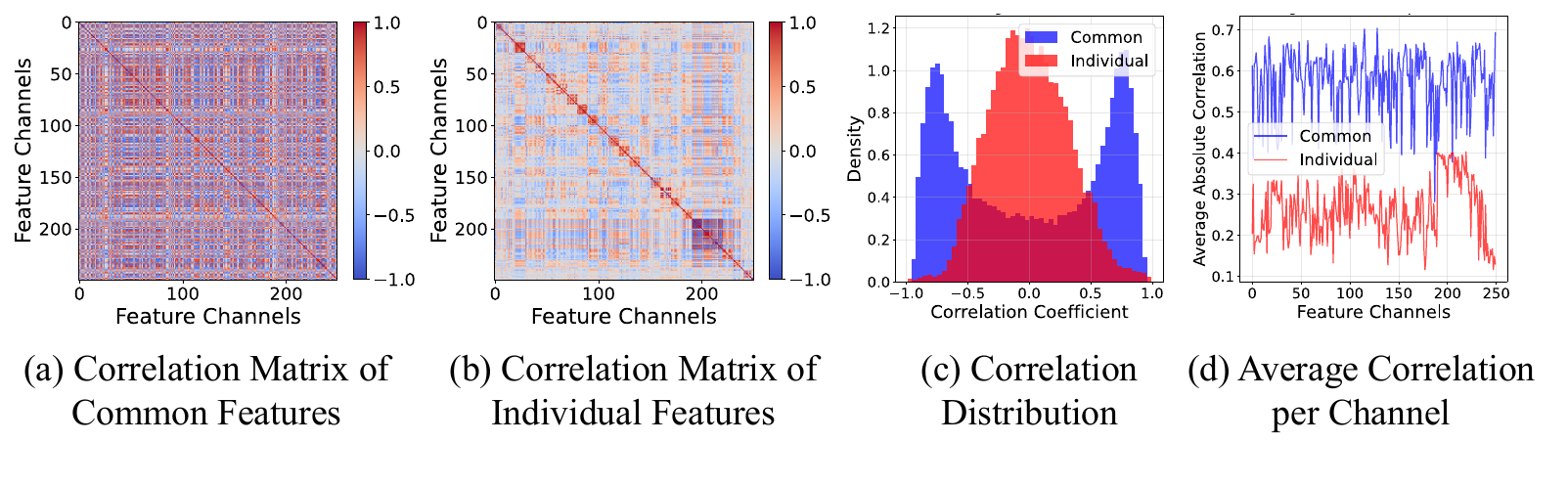}
    \end{center}
    \caption{Qualitative analysis of band split (BS).}
    \label{fig:band_split}
\end{figure}

\noindent\textbf{Effect of Band Split (BS).} As shown in Table~\ref{tab:ablation} (a), BS improves performance (+1.69\% $F_1$, +1.55\% mIoU on HSS) by preserving spectral signatures, confirming that spectral pollution indeed harms segmentation performance. To verify that BS successfully prevents spectral pollution, we compute inter-channel correlations. Figure~\ref{fig:band_split} validates this through correlation analysis: (i) In the common branch without BS, adjacent channels exhibit high correlation coefficients, indicating that features from different bands are mixed during shared encoding, resulting in spectral pollution. (ii) After BS, correlation coefficients between channels belonging to different bands approach zero, while channels within the same band maintain reasonable correlations, demonstrating successful spectral isolation. (iii) The correlation distributions and per-channel analysis further confirm that BS effectively maintains spectral purity for each band.

\begin{figure}[t]
    \begin{center}
        \includegraphics[width=\linewidth]{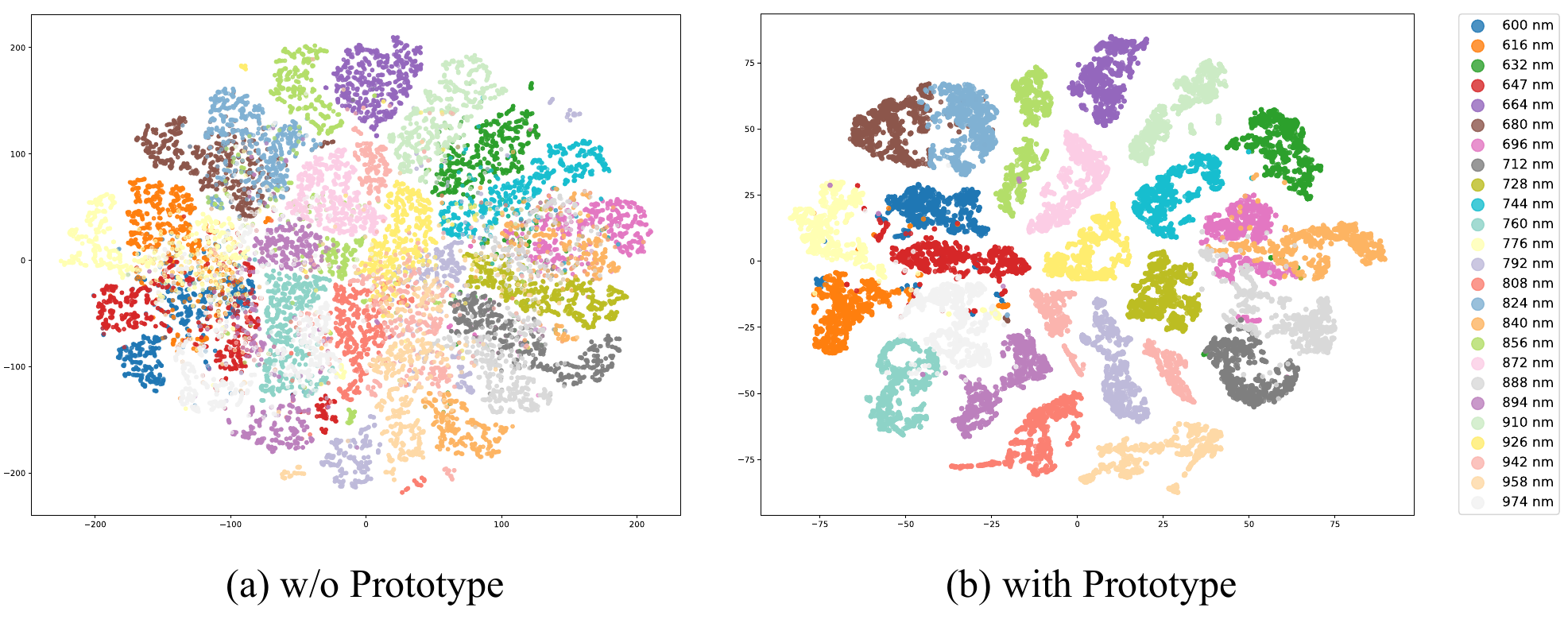}
    \end{center}
    \caption{t-SNE visualization. The PSR produces more clustered and discriminative features than baselines.}
    \label{fig:tsne}
\end{figure}

\noindent\textbf{Effect of Prototype-based Spectral Representation (PSR).} As reported in Table~\ref{tab:ablation} (b), adding PSR achieves cumulative improvements of 71.11\% $F_1$ and 56.10\% mIoU on HSSDataset (+2.82\% $F_1$, +2.94\% mIoU over baseline), with consistent gains on MSSDataset (+1.44\% $F_1$, +1.74\% mIoU), demonstrating that the performance improvement from PSR is consistent across datasets. By learning class prototypes within each band, PSR aggregates discrete per-pixel spectral features into compact representative representations, effectively suppressing intra-class spectral variability while enhancing inter-class separability between smoke and background. Figure~\ref{fig:tsne} provides t-SNE visualization of the learned feature space, showing that with PSR, the feature clusters for smoke and background become more compact with reduced intra-class scatter, and the decision boundary between the two classes becomes clearer with increased inter-class distance.

\begin{figure}[t]
    \begin{center}
        \includegraphics[width=0.90\textwidth]{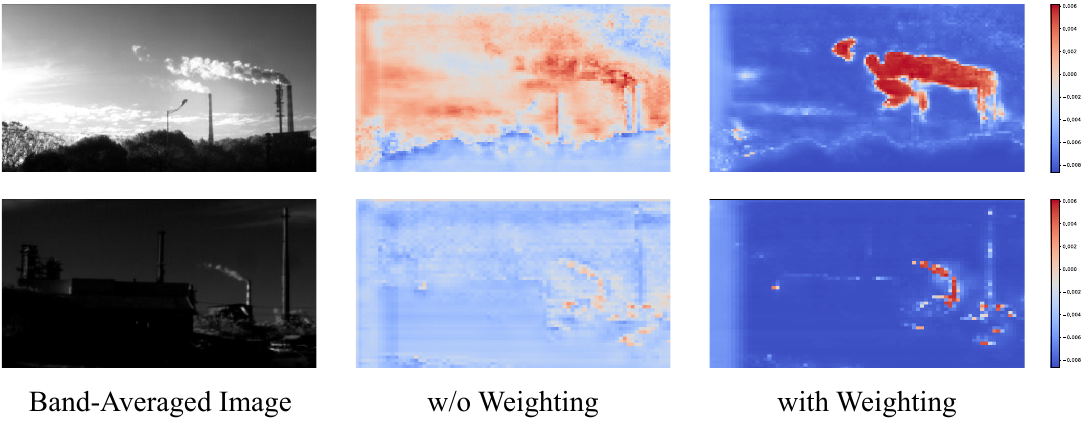}
    \end{center}
    \caption{Effect of the DSR visualized through feature map mean activations.}
    \label{fig:band_weight_overall}
\end{figure}

\noindent\textbf{Effect of Dual-Stage Router (DSR).} We analyze the two components in order. \textit{Feature Router (F-Router):} Compared to BS alone (a$\rightarrow$c), adding the F-Router brings +1.58\% $F_1$ and +1.90\% mIoU, indicating that spatial-aware band weighting is a crucial component for leveraging varying band discriminability. \textit{Prototype Router (P-Router):} On top of PSR + F-Router (d$\rightarrow$e), introducing the P-Router yields an additional +0.72\% $F_1$ and +0.75\% mIoU, confirming that prototype aggregation before band weighting provides complementary gains.
As illustrated in Figure~\ref{fig:band_weight_overall}, we visualize the mean values of feature maps to compare the effects of DSR band weighting versus without band weighting. For interference scenarios (first row), the band weighting mechanism enables the model to produce higher mean activation values in smoke regions while effectively suppressing background interference from clouds and sky. For small smoke in low-light conditions (second row), the weighted feature maps show significantly enhanced mean values in smoke regions, making them more distinguishable from the background.

\noindent\textbf{Component Contribution Analysis.} As evidenced in Table~\ref{tab:ablation}, BS provides the fundamental foundation with +1.69\% $F_1$ improvement, establishing the importance of independent spectral processing. The F-Router shows the largest single-component contribution (+1.58\% $F_1$ when added to BS), highlighting that adaptive spectral weighting is crucial for capturing the varying discriminability across different spectral bands. The combination of PSR and F-Router (d) outperforms either alone (b or c), confirming that prototype learning and band weighting are complementary. The complete MoP (e) outperforms all partial configurations, validating the overall design.

\begin{figure*}[t]
    \begin{center}
        \includegraphics[width=1\linewidth]{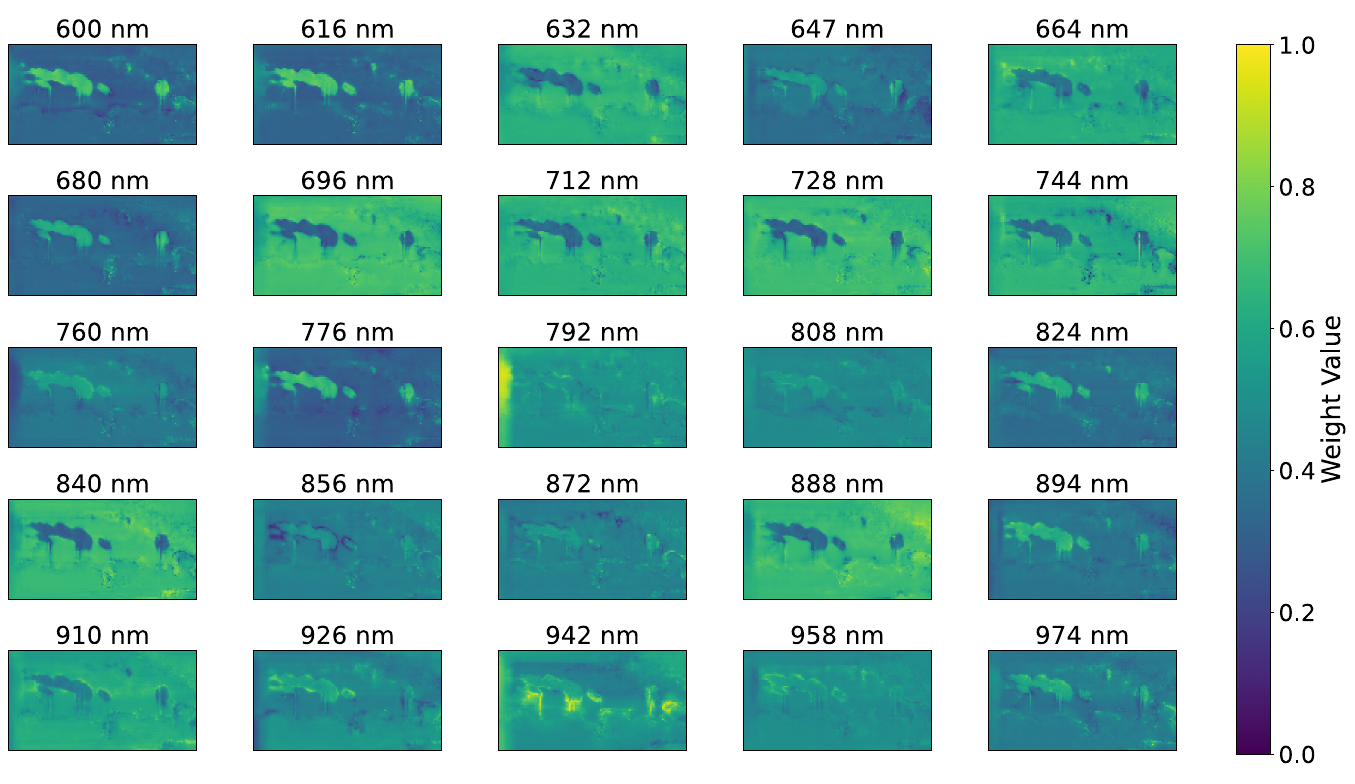}
    \end{center}
    \caption{Visualization of adaptive spectral band weighting learned by the DSR.}
    \label{fig:band_weight}
\end{figure*}

\noindent\textbf{Weighting Distribution.} 
The visualization in Figure~\ref{fig:band_weight} demonstrates the distribution of weighting across different spectral bands:

(i) The band weight distributions vary significantly across different spatial locations. For example, certain bands (e.g., 760nm, 808nm) receive higher weights in smoke body regions, while different bands (e.g., 856nm, 888nm) receive higher weights in semi-transparent edge regions. This is consistent with our core hypothesis that different bands have varying discriminative capabilities in different spatial regions.

(ii) The learned band weight distributions also differ across different scenarios (cloud interference, transparent smoke, small-scale smoke), indicating that the DSR can adaptively adjust band weighting strategies based on specific input scenes.

(iii) Analysis across multiple samples reveals that certain bands tend to receive high weights in specific types of scenarios. For instance, the 942nm band frequently receives high weights when processing smoke emission columns, indicating its sensitivity to specific scene elements.

\subsection{Other Ablation Studies}\label{sec:other_ablation}

\begin{table}[h]
    \centering
    \caption{Analysis of key hyperparameters.}
    \label{tab:hyperparameter_sensitivity}
    \begin{tabular}{l|c|cc}
    \toprule[1pt]
    Parameter & Value & $F_1$ & $mIoU$ \\ \midrule
    \multirow{5}{*}{Prototype Number ($K$)} & 1 & 71.85 & 57.12 \\
    & 2 & 72.94 & 58.31 \\
    & \cellcolor{Gray}3 & \cellcolor{Gray}\textbf{73.63} & \cellcolor{Gray}\textbf{59.21} \\
    & 4 & 73.21 & 58.87 \\
    & 5 & 73.16 & 58.63 \\ \midrule
    \multirow{7}{*}{Feature Dimension ($D$)} & 50 & 71.92 & 57.68 \\
    & 100 & 72.84 & 58.52 \\
    & 150 & 73.15 & 58.76 \\
    & 200 & 73.27 & 58.91 \\
    & \cellcolor{Gray}250 & \cellcolor{Gray}\textbf{73.63} & \cellcolor{Gray}\textbf{59.21} \\
    & 300 & 73.42 & 59.05 \\
    & 500 & 73.18 & 58.87 \\
    \bottomrule[1pt]
    \end{tabular}
\end{table}

\noindent\textbf{The Number of Prototypes.} The number of prototypes $K$ significantly impacts performance (Table~\ref{tab:hyperparameter_sensitivity}). Performance improves progressively as $K$ increases from 1 to 3, indicating that the multi-prototype structure better captures diverse spectral patterns. With $K=3$, our method achieves optimal performance (73.63\% F1, 59.21\% mIoU). When $K$ continues to increase to 4 and 5, performance slightly decreases, likely because MoP independently maintains $K$ prototypes for each class at each band, causing the total number of prototypes to grow rapidly with $K$, reducing the number of pixel samples allocated to each prototype and degrading prototype stability.

\noindent\textbf{The Feature Dimension.} The feature dimension $D$ determines the representational capacity of our prototype features. With $D=250$, our method achieves the best performance. Lower dimensions ($D=50, 100$) lack sufficient representational power, while higher dimensions ($D=500$) may lead to overfitting. The performance variation within $D \in [150, 300]$ is relatively small ($F_1$ between 73.15\% and 73.63\%), indicating model robustness to this hyperparameter.

\noindent\textbf{Prototype Update Strategy.} We compare different prototype update strategies in Table~\ref{tab:prototype_update}. The gradient-based approach, while providing direct optimization through backpropagation, suffers from training instability and overfitting. In contrast, the momentum-based update achieves better performance (73.63\% vs 72.73\% $F_1$) by balancing stability and adaptability.

\begin{table}[!h]
    \centering
    \caption{Ablation study of prototype update strategies.}
    \label{tab:prototype_update}
    \begin{tabular}{l|cc|cc}
    \toprule[1pt]
    \multirow{2}{*}{{Method}} & \multicolumn{2}{c|}{{HSSDataset}} & \multicolumn{2}{c}{{MSSDataset}} \\
    \cmidrule(lr){2-3}\cmidrule(lr){4-5}
     & $F_1$ & $mIoU$ & $F_1$ & $mIoU$ \\ \midrule
    Gradient & 72.73 & 58.08 & 87.78 & 78.41 \\
    \cellcolor{Gray}Momentum & \cellcolor{Gray}\textbf{73.63} & \cellcolor{Gray}\textbf{59.21} & \cellcolor{Gray}\textbf{88.11} & \cellcolor{Gray}\textbf{79.55} \\
    \bottomrule[1pt]
    \end{tabular}
\end{table}

\noindent\textbf{Fusion Strategy.}
Table~\ref{tab:fusion_ablation} compares different ways to combine the common and individual branches. The individual branch alone ($F_1$: 71.89\%) notably outperforms the common branch alone ($F_1$: 70.12\%), confirming that preserving band independence is more critical for discriminability. Element-wise sum achieves the best performance compared to concatenation (HSS: +0.78\% $F_1$, +1.76\% $mIoU$), indicating that this fusion strategy enhances feature integration. The complementary fusion further boosts performance to 73.63\%, validating the synergy between global shared features and band-specific features.

\begin{table}[!h]
    \centering
    \caption{Ablation study of feature fusion strategies.}
    \label{tab:fusion_ablation}
    \begin{tabular}{l|cc|cc}
    \toprule[1pt]
    \multirow{2}{*}{{Fusion Strategy}} & \multicolumn{2}{c|}{{HSSDataset}} & \multicolumn{2}{c}{{MSSDataset}} \\
    \cmidrule(lr){2-3}\cmidrule(lr){4-5}
     & $F_1$ & $mIoU$ & $F_1$ & $mIoU$ \\ \midrule
    Common Only & 70.12 & 54.98 & 85.34 & 76.12 \\
    Spectral Only & 71.89 & 56.23 & 86.67 & 77.89 \\
    Concatenation & 72.85 & 57.45 & 87.23 & 78.34 \\
    \cellcolor{Gray}Element-wise Sum & \cellcolor{Gray}\textbf{73.63} & \cellcolor{Gray}\textbf{59.21} & \cellcolor{Gray}\textbf{88.11} & \cellcolor{Gray}\textbf{79.55} \\
    \bottomrule[1pt]
    \end{tabular}
\end{table}

\section{Conclusion}\label{sec6}

\noindent\textbf{Conclusion.} 
In this work, we introduce the first hyperspectral smoke segmentation task to address the limitations of visible-light-based data in handling cloud interference and semi-transparent smoke regions. We establish the HSSDataset with high-quality annotations across diverse scenarios using a rigorous Many-to-One protocol. To tackle varying discriminative capabilities of spectral bands, we propose a mixture of prototypes (MoP) network with three key components: BS for spectral isolation, PSR for diverse patterns, and DSR for adaptive weighting. We further validate our approach on a multispectral dataset (MSSDataset), demonstrating superior performance across both hyperspectral and multispectral settings. We hope this introduced hyperspectral imaging opens new possibilities for developing more accurate and reliable smoke segmentation systems that can provide earlier warnings and reduce false alarms, thereby improving emergency response and protecting lives and property.

\noindent\textbf{Limitation.} While our method provides an effective strategy for hyperspectral smoke segmentation, our approach employs relatively simple methods in certain components. Specifically, our BS strategy relies on basic group convolution for spectral isolation, and the dual-branch fusion mechanism uses straightforward element-wise averaging. Future work could explore more sophisticated BS techniques and advanced fusion strategies to achieve further performance improvements.

\backmatter

\bibliography{hss}

\end{document}